\documentclass[11pt]{article}

\usepackage[final]{ACL2023}

\usepackage{times}
\usepackage{latexsym}
\usepackage{verbatim}
\usepackage{fancyvrb}    
\usepackage{tcolorbox}
\usepackage{placeins}
\usepackage{xcolor}
\definecolor{olivegreen}{RGB}{85,107,47}
\usepackage{colortbl} 
\usepackage{graphicx}
\usepackage{xspace}
\usepackage{amssymb}
\usepackage{pifont}

\usepackage{tabularx}
\usepackage{booktabs}
\usepackage{float}
\usepackage{multirow}
\usepackage{pgf-pie}
\usepackage{tikz}
\usepackage{caption}
\usepackage{subcaption}
\usepackage[utf8]{inputenc} 
\usepackage[T1]{fontenc}    
\usepackage{hyperref}       
\usepackage{url}            
\usepackage{amsfonts}       
\usepackage{nicefrac}       
\usepackage{microtype}      
\usepackage{xcolor}         
\usepackage{inconsolata}    
\usepackage{stfloats}
\usepackage{paralist}
\usepackage[shortlabels]{enumitem}
\usepackage{listings}

\definecolor{lightestgray}{gray}{0.95}
\newcommand{\varred}[1]{\textcolor{red}{\textbf{#1}}}
\newcommand{\varblue}[1]{\textcolor{blue}{\textbf{#1}}}
\newcommand{\vargreen}[1]{\textcolor{olivegreen}{\textbf{#1}}}

\newcommand{\datasetName}{{\sc MMTabReal}\xspace}
\newcommand{\paperTitle}{\datasetName: Real-World Benchmark for Multimodal Table Understanding}

\tcbuselibrary{listings,breakable} 
\usepackage{enumitem}

\title{\paperTitle}

\author{Prasham Titiya$^\textbf{*}$\quad  Jainil Trivedi$^\textbf{*}$ \quad Chitta Baral \quad Vivek Gupta\textsuperscript{\textdagger} \\
        Arizona State University\\
        \texttt{\{ptitiya,jtrived7,cbaral,vgupt140\}}@asu.edu
        }

\begin{document}
\maketitle
\begingroup
\def\thefootnote{}
\NoHyper
\footnotetext{$\textbf{*}$Equal Contributor \quad \quad \textdagger Primary Superviser}
\endNoHyper
\endgroup\begin{abstract}
Multimodal tables i.e. tabular layouts interleaved with charts, maps, icons, and color encodings are ubiquitous in real applications yet remain difficult for Multimodal Large Language Models (MLLMs). Despite advances in text and image understanding, systematic evaluation of table-centric multimodal reasoning is limited. We introduce \datasetName{}, a MultiModal Table Benchmark, human-curated suite of 500 real-world tables paired with 4{,}021 question–answer pairs. \datasetName{} spans four question types, five reasoning categories, and eight structural archetypes. Evaluations of state-of-the-art models reveal substantial gaps, especially in visual grounding, spatial alignment, and multi-step inference, with 20–40\% performance drops relative to existing benchmarks. These results highlight the need for architectures that more tightly fuse vision with tabular structure and support explicit numeric/logical operations. \datasetName{} is released for evaluation only, providing a rigorous, reproducible testbed that reflects the linguistic, structural, and reasoning complexity of real-world multimodal tables. Code and data are available at:\url{https://coral-lab-asu.github.io/mmtabreal}

\end{abstract}

\section{Introduction}

\begin{figure}[h]
    \centering
    \includegraphics[width=1\linewidth]{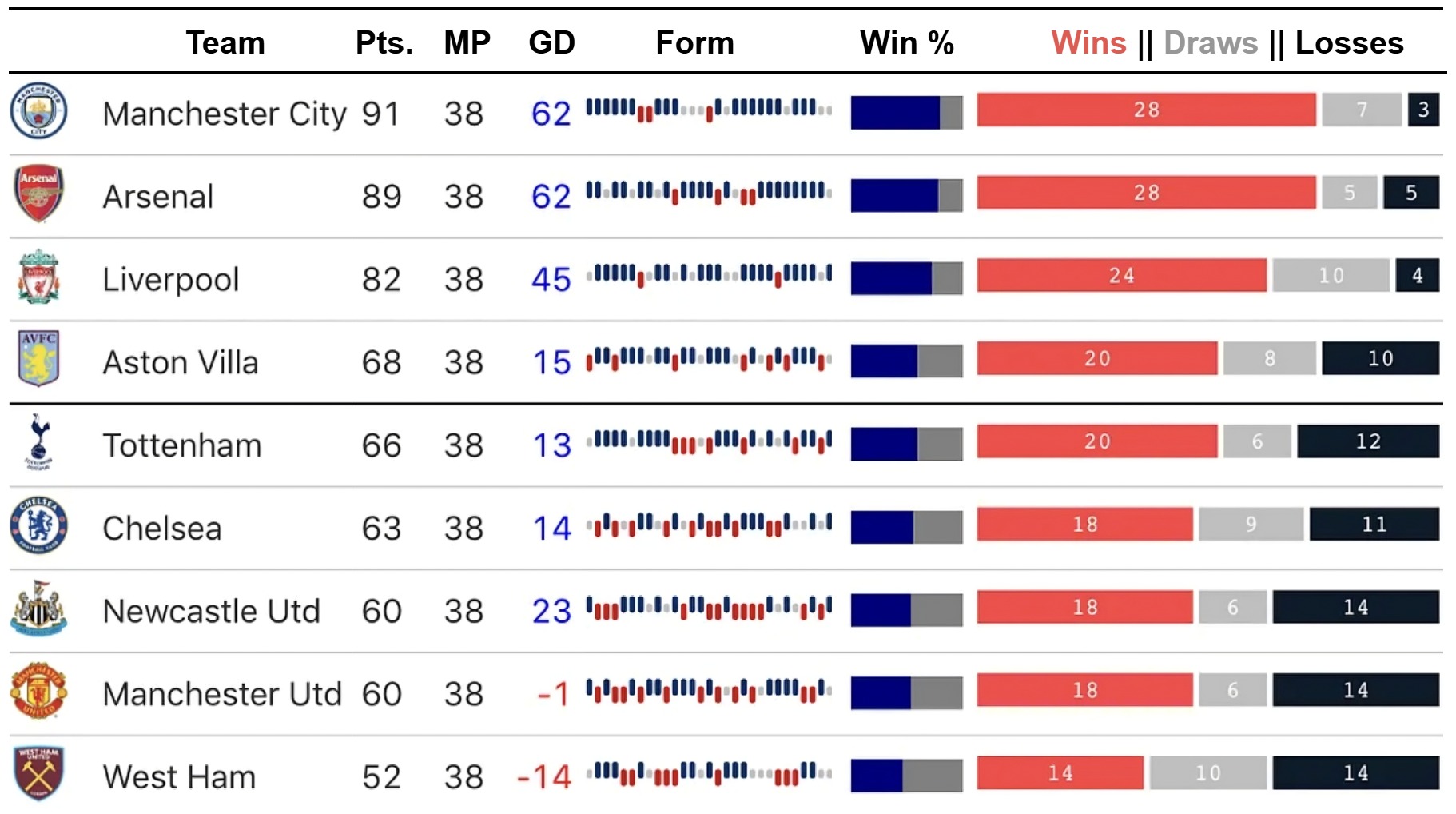}
    {\footnotesize
    \begin{enumerate}[nosep]
    \item[Q1:] Which team has the least number of draws? \\ A1: Arsenal F.C. 
    \item[Q2:] What is the color of the badge of the club with 60 points and a negative goal difference? A2: Red
    \item[Q3:] What is Chelsea's longest winning streak? A3: 3
    \item[Q4:] Which club has the lowest goal difference for a team with a win percentage greater than 50? A4: Tottenham Hotspurs
    \end{enumerate}}
    \vspace{-0.5em}
    \caption{\small A table of English Premier League standings along with accompanying questions and their respective answers}
    \label{fig:main_example_figure}
    \vspace{-1.0em}
\end{figure}

Modern AI is shifting from unimodal perception to structure-aware multimodality, where models jointly read and reason over images embedded in tabular data in both structured and semi-structured \cite{lee2023multimodality,cloutier2021using}. As systems move beyond single-channel inputs, seamless cross-modal integration becomes essential for human-level understanding across domains such as healthcare, finance, and education \cite{10.1145/3536221.3557030}.

Tables are a cornerstone of information representation, providing a two-dimensional scaffold for complex data in both structured and semi-structured forms \cite{SHWARTZZIV202284, jiang2025representation}. Modern tables are no longer text-only: they embed charts, images, color encodings, icons, and logos \cite{zheng2024multimodal}. Unlike fixed-schema databases, this intuitive visual layout is easy for humans to parse but introduces distinct reasoning challenges: (i) header–cell hierarchies are implicit and must be inferred, and (ii) the semantics of adjacent cells are context-dependent \cite{shigarov2023table}.

Reasoning over such data, where textual, numerical, and visual cues are interwoven within a structured layout, is inherently challenging. 
For example, consider the question from Figure~\ref{fig:main_example_figure}: 
\emph{“Which club has the lowest goal difference among teams with a win percentage greater than 50\%?”} 
Answering this requires multi-step, cross-modal reasoning:
\begin{inparaenum}[(1)]
    \item detecting the visual bars in the \textit{Win \%} column to identify clubs exceeding 50\%;
    \item comparing numerical values in the \textit{GD} (goal difference) column;
    \item recognizing club badges, color encodings, and logos for correct team identification;
    \item aligning these visual cues with the textual headers and cell contents to filter candidates; and
    \item concluding that \emph{Tottenham Hotspur} satisfies the condition with the lowest goal difference among the qualifying teams.
\end{inparaenum}
Despite their ubiquity, multimodal tables remain underexplored in AI: most work targets generic vision-language or table-only tasks and overlooks the unique, structure-aware challenges of mixed-format tables \cite{vaishnav2025cognitive}. 
This gap motivates a central question: \emph{Can today’s AI systems, particularly multimodal large language models (MLLMs), effectively reason over complex multimodal tables?}

To address this question, we introduce \datasetName{}, a comprehensive, human-curated benchmark for question answering over \emph{multimodal} tables. Unlike prior resources such as MMTabQA \cite{mathur-etal-2024-knowledge} and MMTab \cite{zheng2024multimodal}, which often rely on basic visuals or synthetic setups, it \datasetName{} spans multimodal web tables with rich, compositional cues: icons, logos, color encodings, and miniature charts paired with naturally phrased questions.

Built from real-world tables and human-authored questions, it captures natural linguistic variation and practical reasoning challenges. \emph{\datasetName{} is evaluation-only, not for training.} It probes whether models can fuse visual and tabular cues, align spatial structure, and perform numerical and logical reasoning, enabling rigorous assessment of MLLMs beyond unimodal capabilities. Our contributions are as follows:
\vspace{0.3em}

\noindent\textbullet\ A human-curated benchmark, \datasetName, for multimodal table reasoning with real-world tables spanning \textbf{8} structural archetypes and \textbf{4K} naturally phrased QA pairs, featuring interleaved text, icons/logos, color encodings, and mini-charts.\\
\noindent\textbullet\  A standardized benchmarking of strong MLLMs shows \textbf{10--30\%} performance drops relative to existing benchmarks across all baselines.\\
\noindent\textbullet\ Fine-grained analyses across \textbf{table types}, \textbf{question types}, \textbf{reasoning skills}, and \textbf{answer formats} reveal consistent models failure modes and actionable patterns.

\section{Motivation} \label{sec:motivation}

\paragraph{Why a new Multimodal TableQA benchmark?}  Existing tableQA datasets are limited in capturing the multimodal complexity of real-world data. MMTabQA \cite{mathur-etal-2024-knowledge} and MMTab \cite{zheng2024multimodal} include basic visual elements like flags, other datasets such as MultimodalQA \cite{talmor2021multimodalqa}, UniMMQA \cite{luo-etal-2023-unifying}, and SPIQA \cite{yvinec2023spiq} offer limited structural or visual diversity. Tables~\ref{tab:multimodal_structure} and ~\ref{tab:multimodal_visuals} summarize these gaps.

\begin{table}[!h]
\vspace{-0.5em}
\scriptsize
\centering
\setlength{\aboverulesep}{0pt}
\setlength{\belowrulesep}{0pt}
\setlength{\tabcolsep}{1pt}
\begin{tabular}{lccccc}
\toprule
\textbf{Dataset} & \textbf{\#Test} & \textbf{Multimodal} & \textbf{Interleaved} & \textbf{TableImage} & \textbf{Hierarchical} \\
\midrule
MMTabQA  & 2,800 & \textcolor{olivegreen}{\checkmark} & \textcolor{olivegreen}{\checkmark} & \textcolor{olivegreen}{\checkmark} & \textcolor{red}{\ding{55}} \\
MMTab  & 49,000 & \textcolor{olivegreen}{\checkmark} & \textcolor{red}{\ding{55}} & \textcolor{olivegreen}{\checkmark} & \textcolor{olivegreen}{\checkmark} \\
MultimodalQA & 3,660 & \textcolor{olivegreen}{\checkmark} & \textcolor{red}{\ding{55}} & \textcolor{red}{\ding{55}} & \textcolor{red}{\ding{55}} \\
UniMMQA  & 4,250 & \textcolor{olivegreen}{\checkmark} & \textcolor{red}{\ding{55}} & \textcolor{red}{\ding{55}} & \textcolor{red}{\ding{55}} \\
SPIQA  & 1,387 & \textcolor{olivegreen}{\checkmark} & \textcolor{red}{\ding{55}} & \textcolor{red}{\ding{55}} & \textcolor{olivegreen}{\checkmark} \\
\rowcolor{yellow!20}\textbf{\datasetName} & 4,021 & \textcolor{olivegreen}{\checkmark} & \textcolor{olivegreen}{\checkmark} & \textcolor{olivegreen}{\checkmark} & \textcolor{olivegreen}{\checkmark} \\
\bottomrule
\end{tabular}
\vspace{-1.0em}
\caption{\small Comparison of structural features in multimodal table QA datasets.}
\label{tab:multimodal_structure}
\vspace{-1.0em}
\end{table}

\begin{table}[!h]
\vspace{-0.5em}
\scriptsize
\centering
\setlength{\aboverulesep}{0pt}
\setlength{\belowrulesep}{0pt}
\setlength{\tabcolsep}{2.5pt}
\begin{tabular}{lcccccccc}
\toprule
\textbf{Dataset} & \textbf{Chart} & \textbf{Map} & \textbf{Visual} & \textbf{Flag} & \textbf{Char.} & \textbf{Loc.} & \textbf{Logo} & \textbf{Symbol} \\
\midrule
MMTabQA & \textcolor{red}{\ding{55}} & \textcolor{red}{\ding{55}} & \textcolor{red}{\ding{55}} & \textcolor{olivegreen}{\checkmark} & \textcolor{olivegreen}{\checkmark} & \textcolor{red}{\ding{55}} & \textcolor{olivegreen}{\checkmark} & \textcolor{olivegreen}{\checkmark} \\
MMTab & \textcolor{red}{\ding{55}} & \textcolor{red}{\ding{55}} & \textcolor{red}{\ding{55}} & \textcolor{olivegreen}{\checkmark} & \textcolor{olivegreen}{\checkmark} & \textcolor{olivegreen}{\checkmark} & \textcolor{olivegreen}{\checkmark} & \textcolor{olivegreen}{\checkmark} \\
MultimodalQA & \textcolor{red}{\ding{55}} & \textcolor{red}{\ding{55}} & \textcolor{red}{\ding{55}} & \textcolor{olivegreen}{\checkmark} & \textcolor{olivegreen}{\checkmark} & \textcolor{red}{\ding{55}} & \textcolor{olivegreen}{\checkmark} & \textcolor{olivegreen}{\checkmark} \\
UniMMQA  & \textcolor{red}{\ding{55}} & \textcolor{red}{\ding{55}} & \textcolor{red}{\ding{55}} & \textcolor{olivegreen}{\checkmark} & \textcolor{olivegreen}{\checkmark} & \textcolor{olivegreen}{\checkmark} & \textcolor{olivegreen}{\checkmark} & \textcolor{olivegreen}{\checkmark} \\
SPIQA & \textcolor{olivegreen}{\checkmark} & \textcolor{red}{\ding{55}} & \textcolor{olivegreen}{\checkmark} & \textcolor{red}{\ding{55}} & \textcolor{red}{\ding{55}} & \textcolor{red}{\ding{55}} & \textcolor{red}{\ding{55}} & \textcolor{red}{\ding{55}} \\
\rowcolor{yellow!20}\textbf{\datasetName} & \textcolor{olivegreen}{\checkmark} & \textcolor{olivegreen}{\checkmark} & \textcolor{olivegreen}{\checkmark} & \textcolor{olivegreen}{\checkmark} & \textcolor{olivegreen}{\checkmark} & \textcolor{olivegreen}{\checkmark} & \textcolor{olivegreen}{\checkmark} & \textcolor{olivegreen}{\checkmark} \\
\bottomrule
\end{tabular}
\vspace{-1.0em}
\caption{\small Comparison of multimodal QA datasets by visual content types.}
\label{tab:multimodal_visuals}
\vspace{-1.0em}
\end{table}

\datasetName provides authentic multimodal tables that combine diverse visual elements with complex structures, requiring true multimodal understanding and enabling effective evaluation of real-world reasoning.

\vspace{0.3em}

\noindent \textbf{Human-Curated vs. Synthetic Benchmarks} Existing multimodal TableQA datasets (MMTabQA \cite{mathur-etal-2024-knowledge}, MMTab \cite{zheng2024multimodal}) are synthetic, hence lacking real-world complexity. Human curation preserves linguistic variation, realistic text–vision relationships, and genuine reasoning, while reducing QA hallucinations \cite{Ji_2023, bang2023multitaskmultilingualmultimodalevaluation} and spurious correlations \cite{zhao2024makesunlearninghard}. Our smaller scale enables stricter quality control. It also better reflects real-world settings in which meaning depends on spatial layout and interdependent visual–textual cues, often lost in pipelines and hard to reproduce synthetically \cite{9712345,lee2023multimodality,cloutier2021using}.

\section{\datasetName Benchmark}
\label{sec:our_dataset}

\datasetName addresses key limitations of synthetic multimodal datasets by focusing on naturally occurring tabular formats and realistic QA needs. 

\subsection{\datasetName Creation}

We built a pipeline to extract high-quality multimodal tables from open-license sources. Automated scripts collected tables, while scanned tables were manually reconstructed to preserve structure and readability. Minor edits removed low-quality images, fixed alignment issues, and standardized formatting. All faces are from public domain sources, and each table includes metadata on structure and visual content.
\vspace{-0.4em}

\paragraph{Annotation Process.} Questions were designed to assess a model’s ability to reason over both textual and visual elements in a table. They were created by NLP experts and cross-reviewed for correctness and consistency by peers and additional reviewers. Every question includes at least one image, either directly or via intermediate reasoning.

We prioritized human annotation over semi-automated methods, which often miss fine-grained visual details and nuanced reasoning and may introduce model-specific biases. This approach ensures challenging, unbiased questions for reliable evaluation.

\noindent
\textbf{Question Types.} We follow the classification scheme of \cite{mathur-etal-2024-knowledge}, grouping questions into four types: (1) \textbf{Explicit Questions} directly reference an entity whose image is present in the table. These help evaluate a model’s ability to directly link textual cues to visual content; (2) \textbf{Implicit Questions} involve an entity whose image is neither explicitly mentioned in the question nor in the answer but plays a crucial role in the intermediate reasoning process. These test a model’s capacity for multi-step reasoning and inference over visual information ; (3) \textbf{Answer-Mention Questions} are characterized by answers containing an entity represented by an image in the table, while the question itself does not explicitly mention this entity. These help assess a model’s ability to retrieve relevant visual entities even when they are not directly queried ; (4) \textbf{Visual-Based Questions} involve tasks that require direct analysis of visual aspects of images, such as color identification, shape recognition, and spatial relationships. They specifically assess a model’s ability to perceive and interpret visual information.

\noindent
\textbf{Reasoning Types.} Questions are categorized by the reasoning they require: (1) \textbf{Extrema Identification} asks for highest or lowest values within the table; (2) \textbf{Mathematical Questions} involve numerical operations and other quantitative computations; (3) \textbf{Fact Verification} questions involve retrieving information from the table to determine whether a given statement is correct; (4) \textbf{Vision-Based} questions involve analyzing visual elements, including shape, patterns; (5) \textbf{Others} encompasses a wide range of reasoning types not covered by other categories, including geography, common-sense and temporal reasoning, and general knowledge tasks.
\subsection{\datasetName Validation}

\datasetName underwent rigorous filtering and verification to ensure quality and reliability. Our process comprised three stages: automated filtering, manual quality control, and annotators validation.
\vspace{-0.3em}
\paragraph{Dataset Filtering:} We implemented comprehensive filtering to remove low-quality content: (i) \textit{Table filtering} removed tables with excessive noise, missing critical data, or formatting issues. This initial process filtered 64 tables; (ii) \textit{Content filtering} eliminated profanity, sensitive personal information, and inappropriate content using manual review; (iii) \textit{Question editing} corrected ambiguous phrasing, unsolvable questions, and logical inconsistencies;

To ensure quality and reliability, we used multi-stage annotation review and validation. Annotators rated correctness on a 3-point scale (0: incorrect, 1: partially correct, 2: correct), with two phases of inter-annotator agreement to assess consistency.

\noindent \textbf{Phase 1 – Internal Review:} Each question was reviewed by two independent annotators to ensure quality. This process corrected 11.26\% of answers and filtered 2.4\% of tables. Table~\ref{tab:answer_agreement} reports Cohen’s Kappa and agreement scores, showing strong annotator agreement across question types.
\begin{table}[htbp]
\vspace{-0.5em}
\centering
\small
\setlength{\aboverulesep}{0pt}
\setlength{\belowrulesep}{0pt}
\setlength{\tabcolsep}{5pt}
\begin{tabular}{lcc}
\toprule
\textbf{Question Type} & \textbf{Cohen's $\kappa$} & \textbf{Percent Agreement} \\
\midrule
Explicit & 0.86 & 96.2\% \\
Implicit & 0.80 & 93.1\% \\
Answer-Mention & 0.88 & 98.0\% \\
Visual & 0.77 & 90.9\% \\
\midrule
\rowcolor{yellow!20}\textbf{Overall Agreement} & \textbf{0.82} & \textbf{96.3\%} \\
\bottomrule
\end{tabular}
\vspace{-0.5em}
\caption{\small Inter-Annotator Agreement on Answer Correctness Evaluation}
\label{tab:answer_agreement}
\vspace{-1.0em}
\end{table}

\noindent \textbf{Phase 2 - External Validation (Subset):} To address potential authorship bias, three external annotators unaffiliated with this research evaluated a subset of 20\% of the questions (804 questions) with no prior knowledge of research objectives or gold standard answers.

Table~\ref{tab:third_reviewer_agreement} presents agreement scores between external annotators and gold answers, confirming the reliability of our work.

\begin{table}[htbp]
\vspace{-0.5em}
\centering
\small
\setlength{\aboverulesep}{0pt}
\setlength{\belowrulesep}{0pt}
\setlength{\tabcolsep}{5pt}
\begin{tabular}{lcc}
\toprule
\textbf{Question Type} & \textbf{Cohen's $\kappa$} & \textbf{Percent Agreement} \\
\midrule
Explicit & 0.89 & 97.0\% \\
Implicit & 0.83 & 94.2\% \\
Answer-Mention & 0.91 & 98.8\% \\
Visual & 0.80 & 91.8\% \\
\midrule
\rowcolor{yellow!20}\textbf{Overall Agreement} & \textbf{0.85} & \textbf{96.6\%} \\
\bottomrule
\end{tabular}
\vspace{-0.5em}
\caption{\small Agreement Between External Reviewers and Gold Standard Evaluation}
\label{tab:third_reviewer_agreement}
\vspace{-1.0em}
\end{table}

Additional details on Scoring and Filtering Guidelines, and Confusion Matrices for both agreements are provided in Appendix~\ref{sec:IAA}.

\subsection{\datasetName Statistics}
\label{sec:dataset_statistics}
\paragraph{Dataset.}
Table~\ref{tab:table_summary} presents key statistics for \datasetName, such as the total number of tables and questions, average structural dimensions, and the proportion of visual elements, illustrating the dataset’s scale and multimodal richness.

\begin{table}[H]
\centering
\small
\setlength{\aboverulesep}{0pt}
\setlength{\belowrulesep}{0pt}
\setlength{\tabcolsep}{5pt}
\begin{tabular}{lc}
\toprule
\textbf{Metric} & \textbf{Value} \\
\midrule
Total Tables & 500 \\
Total Questions & 4,021 \\
Avg. Images per Table & 23.67 \\
Avg. Rows per Table & 19.49 \\
Avg. Columns per Table & 10.85 \\
\% Rows with Images & 89.27 \\
\% Columns with Images & 28.42 \\
\bottomrule
\end{tabular}
\vspace{-0.5em}
\caption{\small \datasetName Dataset Statistics}
\label{tab:table_summary}
\vspace{-1.0em}
\end{table}

\paragraph{Structural and Reasoning Diversity.}
Table~\ref{tab:combined_types} shows \datasetName's diversity across table types, answer types, question types, and reasoning types.

\begin{table}[H]
\vspace{-0.8em}
\centering
\scriptsize

\setlength{\aboverulesep}{2pt}
\setlength{\belowrulesep}{2pt}
\setlength{\tabcolsep}{1pt}
\renewcommand{\arraystretch}{1.25}

\begin{tabular}{l c|l c|l c|l c}

\toprule
\shortstack{\textbf{Table}\\\textbf{Type}} & \shortstack{\textbf{\%}} &
\shortstack{\textbf{Answer}\\\textbf{Type}} & \shortstack{\textbf{\%}} &
\shortstack{\textbf{Question}\\\textbf{Type}} & \shortstack{\textbf{\%}} &
\shortstack{\textbf{Reasoning}\\\textbf{Type}} & \shortstack{\textbf{\%}} \\
\midrule

Single Ent.     & 32.8 & Single Ent.   & 56.5 & Explicit   & 55.5 & Others   & 38.8 \\
Multi Ent.      & 26.6 & Single Num.   & 23.9 & Implicit   & 18.4 & Extrema  & 23.7 \\
Single Chart    & 10.8 & Multi Ents.   & 13.5 & Visual     & 17.1 & Math     & 23.0 \\
Visualization   & 9.6  & Multi Types   & 3.6  & Ans. Men.     & 9.1  & Vision   & 11.0 \\
Ent.+Maps       & 9.6  & Multi Nums.   & 1.9  &            &      & Factual  & 3.5 \\
Ent.+Charts     & 4.2  & Img Loc.      & 0.6  &            &      &          &      \\
Multi Charts    & 4.0  &               &      &            &      &          &      \\
Maps Only       & 2.4  &               &      &            &      &          &      \\
\midrule

\textbf{Total} & \textbf{100.0} &
\textbf{Total} & \textbf{100.0} &
\textbf{Total} & \textbf{100.0} &
\textbf{Total} & \textbf{100.0} \\
\bottomrule

\end{tabular}

\vspace{-0.6em}
\caption{\small Distribution of Table Type, Answer Type, Question Type, and Reasoning Type. Ent.=Entity, Num.=Number, Chart.=Chart, Viz.=Visualization, Ans. Men.=Answer-Mention, Extrema=Finding extrema values, Math=Mathematical reasoning, Vision=Vision-based reasoning.}
\label{tab:combined_types}
\vspace{-1.0em}
\end{table}

\begin{table}[htbp]
\vspace{-0.5em}
\centering
\scriptsize
\setlength{\aboverulesep}{0pt}
\setlength{\belowrulesep}{0pt}
\setlength{\tabcolsep}{4pt}
\begin{tabular}{lc|lc}
    \toprule
    \textbf{Image Type} & \textbf{\% of Total} &\textbf{Image Type} & \textbf{\% of Total}\\
    \midrule
    Human / Fictional Character      & 21.85 & Chart                            &  6.85 \\
    Flag / Coat of Arms / Seals      & 18.19 & Location                         &  5.34 \\
    Logo                             & 15.73 &  Symbol                           &  6.19 \\
    Map                              &  9.01 & Poster / Covers                  &  5.76 \\
    Visualizations                   &  2.02 &   Other                            &  9.05 \\
    \midrule
    \textbf{Total}                   & \textbf{11,836} \\
    \bottomrule
\end{tabular}
\vspace{-1.0em}
\caption{\small Image Type Distribution}
\label{tab:image_distribution}
\vspace{-1.0em}
\end{table}
\vspace{-0.5em}

\paragraph{Image Type Distribution.} Table~\ref{tab:image_distribution} shows the distribution of the types of images present in the dataset, demonstrating comprehensive coverage across visual domains relevant to real-world applications.

\subsection{Dataset Examples}

\subsubsection*{Multiple Charts}

\begin{figure}[h]
    \centering
    \includegraphics[width=1.0\linewidth]{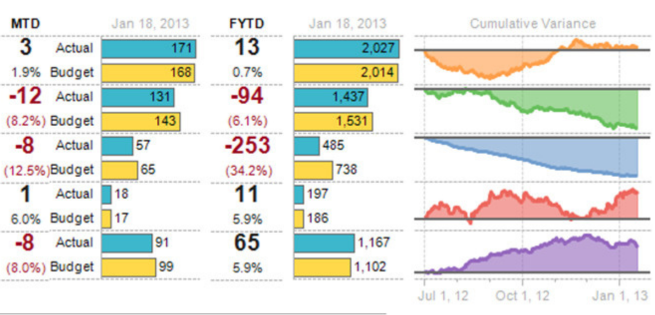}
    {\footnotesize
    \begin{enumerate}[nosep]
    \item[\textbf{Q:}] What is the actual MTD for the third metric on January 18, 2013?
    \textbf{A:} $57$ \quad \textbf{Type:} Explicit.
    \item[\textbf{Q:}] How many FYTD units were sold in total? \\
    \textbf{A:} $5313$ \quad \textbf{Type:} Implicit.
    \item[\textbf{Q:}] Which metric reported the greatest increase? \\
    \textbf{A:} FYTD \quad \textbf{Type:} Answer-Mention.
    \item[\textbf{Q:}] What colored variance graph shows a steady decline over time?
    \textbf{A:} Blue \quad \textbf{Type:} Visual.
    \end{enumerate}}
    \vspace{-0.7em}
    \caption{\small An example of a multimodal table with multiple charts.}
    \label{fig:mult_chart}
    \vspace{-1.0em}
\end{figure}

Figure~\ref{fig:mult_chart} shows a multimodal financial dashboard with MTD and FYTD comparisons of actual vs.\ budget across five metrics, along with cumulative variance trends.

\subsubsection*{Visualizations}

\begin{figure}[h]
\vspace{-0.5em}
    \centering
    \includegraphics[width=0.9\linewidth]{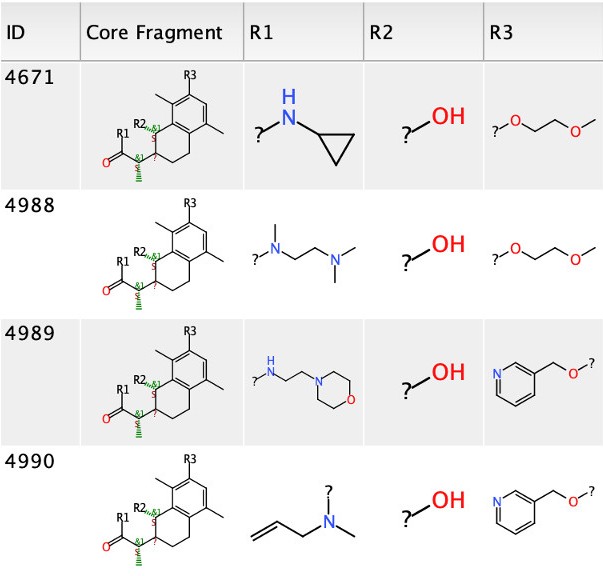}
    {\footnotesize
    \begin{enumerate}[nosep]
    \item[\textbf{Q:}] Which ID numbers have a benzene bases R3? \\
    \textbf{A:} [$4989$, $4990$] \quad \textbf{Type:} Explicit.
    \item[\textbf{Q:}] How many Nitrogen atoms exist in R1 and R3 combined?
    \textbf{A:} $8$ \quad \textbf{Type:} Implicit.
    \item[\textbf{Q:}] What is the name of the common ion in R2? \\
    \textbf{A:} Hydroxide Ion \quad \textbf{Type:} Answer-Mention.
    \item[\textbf{Q:}] How many Hydrogen atoms exist in the R1 of 4988? \\
    \textbf{A:} $13$ \quad \textbf{Type:} Visual.
    \end{enumerate}}
    \vspace{-0.7em}
    \caption{\small An example of a multimodal table with visualizations.}
    \label{fig:viz}
    \vspace{-1.0em}
\end{figure}

Figure~\ref{fig:viz} shows a table of four chemical compounds (by ID) with a shared core fragment and three attachment points. The R1, R2, and R3 columns specify the groups attached at each position. Other dataset examples for different table types can be found in Appendix~\ref{sec:examples}.

\vspace{-0.5em}

\section{Experiments and Analysis}\label{sec:experiments}

\vspace{-0.5em}

Using established multimodal evaluation practices, we benchmark on open and closed-source models.

\vspace{-0.5em}

\subsection{Modeling Strategies}
To evaluate model performance on our dataset, we adopt the five benchmarking strategies proposed by \cite{mathur-etal-2024-knowledge}: \textbf{(1) Missing Image Baseline} establishes a \textbf{lower performance bound} by removing all images from the table. Models must infer missing visual information solely from surrounding text, revealing their ability to reason under partial multimodal input. \textbf{(2) Entity Replaced Baseline} serves as an \textbf{upper performance bound}, manually replacing all images with precise textual descriptions. This setup measures reasoning ability under fully informative conditions, eliminating uncertainty from missing visuals. Image-to-text conversion follows a semi-automatic process: \textbf{entity images} and \textbf{maps} are replaced via Google Reverse Image Search, while \textbf{visualizations/charts} are described by multimodal LLMs. All outputs are human-verified. \textbf{(3) Image Captioning Baseline} transforms the multimodal task into a text-only setting by substituting each image with an automatically generated caption. Multimodal LLMs produce context-specific descriptions that are inserted into the tables, allowing analysis of how well models can extract and convey visual information through text. \textbf{(4) Table-as-Image Baseline} renders the entire table as an image, requiring models to interpret all information visually. We use Selenium to convert HTML tables while preserving structure and content, enabling assessment of models’ ability to parse and reason over visually encoded table. \textbf{(5) Interleaved Baseline} preserves the original multimodal format, keeping images embedded within tables. This setting requires simultaneous reasoning over textual visual information, providing the comprehensive evaluation of models’ capacity for multimodal integration and joint reasoning.

\subsection{LLMs, Prompting, and Metric}

\paragraph{LLMs:} We evaluated multiple models across baselines: Gemini 1.5/2.0 \cite{team2024gemini}, GPT-4o Mini, Llama3-8b \cite{touvron2023llama}, and Mixtral-8x7B \cite{jiang2024mixtral} for text baselines. Vision-capable baselines additionally included InternVL2.5-8B \cite{kweon-etal-2023-open}, Mantis-8B-Idefics2 \cite{jiang2024mantis}, Phi-3.5-Vision \cite{abdin2024phi}, Qwen 2.5-VL \cite{wang2024qwen2}, Qwen 3-VL \cite{bai2025qwen3vltechnicalreport}. Gemma3 27B-it \cite{gemmateam2025gemma3technicalreport} and Table-LLaVA \cite{zheng2024multimodal}. API calls were used for GPT and Gemini 1.5 and 2.0 inference. The remaining models were run locally from HuggingFace Transformers on an A100 GPU with 32 GB of memory.
Parameters include a temperature of 0.2 and a maximum output length of 1024 tokens.

\vspace{-0.3em}
\paragraph{Prompting Strategies:} We used 1-shot prompting for text baselines (Missing Image, Entity Replaced) and 0-shot for image baselines (Image Captioning, Table as Image, Interleaved). 1-shot helps text models by clarifying the task without overwhelming them~\cite{sahoo2025systematicsurveypromptengineering}, while examples can distract multimodal models~\cite{ma-etal-2025-caution}, ensuring fair and optimal evaluation. Full prompts are in Appendix~\ref{sec:prompt}. All tables are provided in the pipe seperated format.
\vspace{-0.3em}
\paragraph{Evaluation Metrics:} We use standard metrics from prior work (Section ~\ref{sec:prior}) adapted to different answer types. \textbf{(a) Exact Match (EM)} checks if the prediction exactly matches the ground truth, used for single-number, single-entity, and image location answers. \textbf{(b) Substring Match (SS)} verifies if the prediction appears within the ground truth, allowing partial matches for multiple entities or numbers. \textbf{(c) F1-Score} computes the harmonic mean of precision and recall, suitable for multiple entities, numbers, and combinations.


\begin{table*}[t]
\vspace{-0.5em}
\scriptsize
\centering
\setlength{\aboverulesep}{0pt}
\setlength{\belowrulesep}{0pt}
\setlength{\tabcolsep}{8pt}
\begin{tabular}{lccc|ccc|ccc|ccc}
\toprule
\multicolumn{1}{c}{} & \multicolumn{3}{c|}{\textbf{Answer Mention}} & \multicolumn{3}{c|}{\textbf{Explicit}} & \multicolumn{3}{c|}{\textbf{Implicit}} & \multicolumn{3}{c}{\textbf{Visual Question}} \\
\cmidrule{2-13}
Model & EM & SS & F1 & EM & SS & F1 & EM & SS & F1 & EM & SS & F1 \\
\cmidrule{2-13}
& \multicolumn{12}{c}{\textbf{Missing Image Baseline}} \\
\cmidrule{1-13}
Gemini 1.5 Flash & 26.59 & 27.39 & 0.128 & 19.52 & 20.92 & 0.085 & 15.32 & 15.29 & 0.063 & 12.91 & 13.84 & 0.054 \\
Gemini 2.0 Flash & 27.98 & 30.15 & 0.089 & 19.31 & 21.03 & 0.075 & 14.12 & 14.77 & 0.052 & 17.60 & 18.51 & 0.064 \\
GPT-4o Mini & 38.99 & 38.40 & 0.294 & 33.97 & 36.24 & 0.251 & 24.14 & 25.71 & 0.143 & 27.00 & 27.33 & 0.163 \\
Llama 3-8B & 32.50 & 32.27 & 0.219 & 29.39 & 28.69 & 0.194 & 22.91 & 23.09 & 0.129 & 20.74 & 20.84 & 0.133 \\
Mixtral & \cellcolor{yellow}42.84 & \cellcolor{yellow}46.31 & \cellcolor{yellow}0.321 & \cellcolor{yellow}36.21 & \cellcolor{yellow}40.70 & \cellcolor{yellow}0.282 & \cellcolor{yellow}28.56 & \cellcolor{yellow}33.46 & \cellcolor{yellow}0.202 & \cellcolor{yellow}30.29 & \cellcolor{yellow}34.48 & \cellcolor{yellow}0.241 \\
\cmidrule{2-13}
& \multicolumn{12}{c}{\textbf{Entity Replaced Baseline}} \\
\cmidrule{2-13}
Gemini 1.5 Flash & 59.89 & 67.20 & 0.394 & 54.71 & 54.61 & 0.295 & 43.73 & 47.16 & 0.238 & - & - & - \\
Gemini 2.0 Flash & 59.50 & 62.46 & 0.293 & 59.93 & 60.04 & 0.300 & 39.71 & 41.26 & 0.177 & - & - & - \\
GPT-4o Mini & \cellcolor{yellow}68.14 & \cellcolor{yellow}70.38 & \cellcolor{yellow}0.538 & \cellcolor{yellow}65.99 & \cellcolor{yellow}69.67 & \cellcolor{yellow}0.496 & \cellcolor{yellow}50.59 & \cellcolor{yellow}52.73 & \cellcolor{yellow}0.340 & - & - & - \\
Llama 3-8B & 61.49 & 62.57 & 0.478 & 54.92 & 57.85 & 0.409 & 41.56 & 44.79 & 0.285 & - & - & - \\
Mixtral & 59.74 & 68.01 & 0.531 & 60.77 & 66.71 & 0.475 & 43.67 & 48.70 & 0.308 & - & - & - \\
\cmidrule{2-13}
& \multicolumn{12}{c}{\textbf{Image Captioning Baseline}} \\
\cmidrule{2-13}
Gemini 1.5 Flash & 29.70 & 30.79 & 0.224 & 30.12 & 32.74 & 0.219 & 18.91 & 19.45 & 0.126 & 21.44 & 24.04 & 0.156 \\
Gemini 2.0 Flash & \cellcolor{yellow}36.82 & \cellcolor{yellow}38.45 & \cellcolor{yellow}0.261 & \cellcolor{yellow}36.82 & \cellcolor{yellow}38.45 & \cellcolor{yellow}0.261 & \cellcolor{yellow}19.69 & \cellcolor{yellow}20.50 & \cellcolor{yellow}0.124 & \cellcolor{yellow}25.09 & \cellcolor{yellow}27.23 & \cellcolor{yellow}0.185 \\
\cmidrule{2-13}
& \multicolumn{12}{c}{\textbf{Table as an Image Baseline}} \\
\cmidrule{2-13}
Gemini 1.5 Flash & 38.39 & 36.22 & 0.178 & 30.16 & 31.30 & 0.148 & 25.14 & 27.52 & 0.113 & 25.66 & 27.80 & 0.103 \\
Gemini 2.0 Flash & 40.44 & 38.98 & 0.212 & 38.55 & 38.18 & 0.214 & 33.83 & 35.92 & 0.199 & 30.49 & 34.05 & 0.195 \\
GPT-4o Mini & \cellcolor{yellow}48.96 & \cellcolor{yellow}50.59 & \cellcolor{yellow}0.357 & \cellcolor{yellow}47.53 & \cellcolor{yellow}49.78 & \cellcolor{yellow}0.345 & \cellcolor{yellow}38.86 & \cellcolor{yellow}40.49 & \cellcolor{yellow}0.265 & \cellcolor{yellow}38.56 & \cellcolor{yellow}41.11 & \cellcolor{yellow}0.291 \\
Intern-VL-2.5 & 19.55 & 40.26 & 0.199 & 18.55 & 38.53 & 0.176 & 16.42 & 36.90 & 0.153 & 14.47 & 38.63 & 0.162 \\
Mantis & 20.85 & 23.23 & 0.109 & 19.72 & 20.90 & 0.113 & 20.88 & 21.49 & 0.110 & 18.60 & 20.26 & 0.107 \\
Phi-3.5 & 21.63 & 23.86 & 0.111 & 18.09 & 19.80 & 0.076 & 15.67 & 16.96 & 0.057 & 17.81 & 19.66 & 0.093 \\
Qwen-2.5-VL & 34.61 & 38.86 & 0.174 & 30.62 & 34.58 & 0.159 & 19.64 & 22.64 & 0.108 & 21.35 & 24.38 & 0.124 \\
Qwen-3-VL & 41.19 & 45.66 & 0.258 & 38.82 & 41.86 & 0.305 & 29.51 & 31.85 & 0.173 & 32.48 & 39.16 & 0.228 \\
Table LLava-1.5-7B & 10.30 & 11.43 & 0.062 & 12.68 & 14.49 & 0.063 & 15.77 & 16.52 & 0.060 & 10.95 & 11.30 & 0.050 \\
\cmidrule{2-13}
& \multicolumn{12}{c}{\textbf{Interleaved Baseline}} \\
\cmidrule{2-13}
Gemini 1.5 Flash & 34.38 & 35.24 & 0.247 & 31.55 & 31.52 & 0.210 & 20.33 & 20.47 & 0.119 & 26.29 & 25.65 & 0.175 \\
Gemini 2.0 Flash & 37.27 & 38.47 & 0.272 & 34.08 & 37.46 & 0.231 & 24.59 & 25.75 & 0.142 & 26.38 & 28.76 & 0.176 \\
GPT-4o Mini & \cellcolor{yellow}\textbf{47.74} & 49.88 & \cellcolor{yellow}\textbf{0.376} & \cellcolor{yellow}\textbf{46.92} & 48.96 & \cellcolor{yellow}\textbf{0.348} & \cellcolor{yellow}\textbf{36.41} & 37.84 & \cellcolor{yellow}\textbf{0.260} & \cellcolor{yellow}\textbf{40.39} & 42.64 & \cellcolor{yellow}\textbf{0.303} \\
Mantis & 24.76 & 26.45 & 0.156 & 24.37 & 26.57 & 0.150 & 24.92 & 26.58 & 0.113 & 20.70 & 23.12 & 0.126 \\
Phi-3.5 & 20.85 & 23.72 & 0.120 & 21.63 & 23.61 & 0.114 & 23.83 & 26.85 & 0.134 & 17.71 & 18.95 & 0.100 \\
Qwen-2.5-VL & 35.66 & \cellcolor{yellow}\textbf{53.45} & 0.271 & 30.35 & \cellcolor{yellow}\textbf{57.02} & 0.258 & 17.95 & \cellcolor{yellow}\textbf{50.59} & 0.146 & 23.04 & \cellcolor{yellow}\textbf{47.94} & 0.200 \\
Qwen-3-VL & 41.19 & 45.66 & 0.358 & 38.82 & 41.86 & 0.305 & 29.51 & 31.85 & 0.173 & 32.48 & 39.16 & 0.294 \\
Gemma3 27B it & 36.49 & 49.16 & 0.340 & 34.49 & 48.77 & 0.313 & 25.99 & 35.08 & 0.175 & 29.03 & 39.08 & 0.207 \\
\cmidrule{2-13}
Human Baseline & \textbf{78.4} & \textbf{82.1} & \textbf{0.76} & \textbf{84.2} & \textbf{87.3} & \textbf{0.81} & \textbf{75.8} & \textbf{80.6} & \textbf{0.73} & \textbf{79.9} & \textbf{83.7} & \textbf{0.78} \\
\bottomrule
\end{tabular}
\vspace{-0.5em}
\caption{\small Performance analysis across question types. EM: Exact Match, SS: Substring Match, F1: F1 Score. Human Baseline denotes a subset of the data.}
\label{tab:question_types_results}
\vspace{-1em}
\end{table*}

\begin{table*}[t]
\vspace{-0.5em}
\scriptsize
\centering
\setlength{\aboverulesep}{0pt}
\setlength{\belowrulesep}{0pt}
\setlength{\tabcolsep}{8.5pt}
\begin{tabular}{lccc|ccc|ccc|ccc}
\toprule
\multicolumn{1}{c}{} & \multicolumn{3}{c|}{\textbf{Single Entity}} & \multicolumn{3}{c|}{\textbf{Multiple Entities}} & \multicolumn{3}{c|}{\textbf{Single Chart}} & \multicolumn{3}{c}{\textbf{Multiple Charts}} \\
\cmidrule{2-13}
Model & EM & SS & F1 & EM & SS & F1 & EM & SS & F1 & EM & SS & F1 \\
\cmidrule{2-13}
& \multicolumn{12}{c}{\textbf{Table as an Image Baseline}} \\
\cmidrule{1-13}
 Gemini 1.5 Flash     & 39.36 & 39.83 & 0.22 & 38.41 & 39.07 & 0.20 & 40.78 & 41.55 & 0.23 & 31.39 & 33.46 & 0.22 \\
 Gemini 2.0 Flash     & 44.42 & 46.37 & 0.27 & 41.63 & 42.65 & 0.25 & 45.38 & 48.33 & 0.30 & 32.92 & 33.92 & 0.19 \\
 GPT-4o Mini          & \cellcolor{yellow}47.49 & \cellcolor{yellow}49.75 & \cellcolor{yellow}0.29 & \cellcolor{yellow}44.99 & \cellcolor{yellow}46.85 & \cellcolor{yellow}0.28 & \cellcolor{yellow}54.96 & \cellcolor{yellow}57.88 & \cellcolor{yellow}0.37 & \cellcolor{yellow}36.43 & \cellcolor{yellow}39.72 & \cellcolor{yellow}0.22 \\
 Intern-VL-2.5        & 33.08 & 45.69 & 0.25 & 35.03 & 53.24 & 0.27 & 34.55 & 56.89 & 0.25 & 35.58 & 50.58 & 0.25 \\
 Mantis-8B-Idefics2   & 30.04 & 30.80 & 0.16 & 32.77 & 34.51 & 0.19 & 44.98 & 46.31 & 0.24 & 37.55 & 47.14 & 0.24 \\
 Phi-3.5              & 31.75 & 32.72 & 0.18 & 26.82 & 28.15 & 0.16 & 36.09 & 37.87 & 0.20 & 31.62 & 32.28 & 0.20 \\
 Qwen-2.5-VL          & 39.65 & 41.64 & 0.23 & 36.53 & 39.51 & 0.23 & 36.58 & 38.96 & 0.21 & 35.63 & 36.81 & 0.23 \\
 Qwen-3-VL            & 43.20 & 45.30 & 0.26 & 40.25 & 42.80 & 0.25 & 48.50 & 51.20 & 0.32 & 35.80 & 38.10 & 0.21 \\
 Table\_LLaVA         & 26.23 & 27.35 & 0.14 & 28.75 & 30.01 & 0.14 & 27.47 & 29.07 & 0.15 & 30.75 & 33.49 & 0.20 \\
\cmidrule{2-13}
& \multicolumn{12}{c}{\textbf{Interleaved Baseline}} \\
\cmidrule{2-13}
 Gemini 1.5 Flash     & 40.89 & 41.72 & 0.22 & 28.43 & 29.53 & 0.17 & 30.48 & 32.52 & 0.18 & 25.51 & 26.01 & 0.14 \\
 Gemini 2.0 Flash     & 40.96 & 42.42 & 0.22 & 32.77 & 33.99 & 0.19 & 40.77 & 43.63 & 0.22 & 33.33 & 34.67 & 0.20 \\
 GPT-4o Mini          & \cellcolor{yellow}45.83 & 47.26 & \cellcolor{yellow}0.28 & \cellcolor{yellow}43.14 & 44.48 & \cellcolor{yellow}0.27 & \cellcolor{yellow}52.99 & 55.61 & \cellcolor{yellow}0.34 & \cellcolor{yellow}38.56 & 41.39 & \cellcolor{yellow}0.27 \\
 Mantis-8B-Idefics2   & 34.80 & 36.59 & 0.19 & 34.24 & 36.19 & 0.18 & 40.66 & 41.27 & 0.22 & 42.38 & 42.99 & 0.23 \\
 Phi-3.5              & 39.75 & 41.65 & 0.21 & 38.28 & 42.32 & 0.24 & 44.23 & 46.71 & 0.25 & 33.28 & 34.08 & 0.20 \\
 Qwen-2.5-VL          & 29.65 & \cellcolor{yellow}61.37 & 0.19 & 29.06 & \cellcolor{yellow}64.65 & 0.18 & 30.10 & \cellcolor{yellow}76.22 & 0.18 & 30.53 & \cellcolor{yellow}59.01 & 0.17 \\
 Qwen-3-VL            & 42.50 & 44.20 & 0.25 & 39.80 & 41.50 & 0.24 & 47.20 & 49.80 & 0.30 & 35.30 & 37.60 & 0.24 \\
\midrule
 Model & \multicolumn{3}{c|}{\textbf{Maps Only}} & \multicolumn{3}{c|}{\textbf{Visualizations}} & \multicolumn{3}{c|}{\textbf{Entities \& Maps}} & \multicolumn{3}{c}{\textbf{Entities \& Charts}} \\
\cmidrule{2-13}
& \multicolumn{12}{c}{\textbf{Table as an Image Baseline}} \\
\cmidrule{1-13}
 Gemini 1.5 Flash     & 25.44 & 26.50 & 0.15 & 50.57 & 51.60 & 0.27 & 35.23 & 36.30 & 0.22 & 22.63 & 23.65 & 0.20 \\
 Gemini 2.0 Flash     & \cellcolor{yellow}28.27 & \cellcolor{yellow}29.30 & \cellcolor{yellow}0.18 & \cellcolor{yellow}55.56 & \cellcolor{yellow}56.60 & \cellcolor{yellow}0.30 & 38.66 & 39.70 & 0.25 & \cellcolor{yellow}24.64 & \cellcolor{yellow}25.65 & \cellcolor{yellow}0.22 \\
 GPT-4o Mini          & 27.12 & 28.15 & 0.17 & 54.40 & 55.45 & 0.29 & \cellcolor{yellow}40.03 & \cellcolor{yellow}41.05 & \cellcolor{yellow}0.27 & 23.35 & 24.38 & 0.21 \\
 Intern-VL-2.5        & 15.42 & 16.45 & 0.09 & 30.68 & 31.70 & 0.17 & 20.23 & 21.20 & 0.14 & 12.88 & 13.89 & 0.09 \\
 Mantis-8B-Idefics2   & 15.51 & 16.50 & 0.10 & 35.30 & 36.25 & 0.18 & 22.27 & 23.25 & 0.16 & 14.98 & 15.95 & 0.10 \\
 Phi-3.5              & 15.95 & 16.90 & 0.09 & 32.00 & 33.00 & 0.17 & 18.58 & 19.50 & 0.13 & 12.58 & 13.50 & 0.08 \\
 Qwen-2.5-VL          & 26.87 & 27.85 & 0.16 & 52.85 & 53.90 & 0.26 & 37.63 & 38.60 & 0.23 & 22.36 & 23.35 & 0.19 \\
 Qwen-3-VL            & 27.50 & 28.50 & 0.17 & 54.20 & 55.30 & 0.28 & 39.40 & 40.40 & 0.26 & 23.80 & 24.80 & 0.21 \\
 Table\_LLaVA         & 12.28 & 13.28 & 0.08 & 28.71 & 29.75 & 0.15 & 19.73 & 20.70 & 0.12 & 10.84 & 11.85 & 0.07 \\
\cmidrule{2-13}
& \multicolumn{12}{c}{\textbf{Interleaved Baseline}} \\
\cmidrule{2-13}
 Gemini 1.5 Flash     & 23.44 & 24.50 & 0.13 & 48.57 & 49.60 & 0.25 & 33.23 & 34.30 & 0.20 & 20.63 & 21.65 & 0.18 \\
 Gemini 2.0 Flash     & \cellcolor{yellow}26.27 & \cellcolor{yellow}27.30 & \cellcolor{yellow}0.16 & \cellcolor{yellow}53.56 & \cellcolor{yellow}54.60 & \cellcolor{yellow}0.28 & 36.66 & 37.70 & 0.23 & \cellcolor{yellow}22.64 & \cellcolor{yellow}23.65 & \cellcolor{yellow}0.20 \\
 GPT-4o Mini          & 25.12 & 26.15 & 0.15 & 52.40 & 53.45 & 0.27 & \cellcolor{yellow}38.03 & \cellcolor{yellow}39.05 & \cellcolor{yellow}0.25 & 21.35 & 22.38 & 0.19 \\
 Mantis-8B-Idefics2   & 13.51 & 14.50 & 0.08 & 33.30 & 34.25 & 0.16 & 20.27 & 21.25 & 0.14 & 12.98 & 13.95 & 0.08 \\
 Phi-3.5              & 13.95 & 14.90 & 0.07 & 30.00 & 31.00 & 0.15 & 16.58 & 17.50 & 0.11 & 10.58 & 11.50 & 0.06 \\
 Qwen-2.5-VL          & 24.87 & 25.85 & 0.14 & 50.85 & 51.90 & 0.24 & 35.63 & 36.60 & 0.21 & 20.36 & 21.35 & 0.17 \\
 Qwen-3-VL            & 25.40 & 26.40 & 0.15 & 52.10 & 53.20 & 0.26 & 37.20 & 38.20 & 0.24 & 21.60 & 22.60 & 0.19 \\
\bottomrule
\end{tabular}
\vspace{-0.5em}
\caption{\small Performance analysis across table types. EM: Exact Match, SS: Substring Match, F1: F1 Score}
\label{tab:table_types}
\vspace{-1em}
\end{table*}


\subsection{Results and Analysis}

\textbf{Across modeling strategies:}\
Missing Images performs worst, highlighting the importance of visual input. Image Captioning shows the smallest gains, likely due to its query-independent, static descriptions that miss fine-grained details (e.g., small icons, chart scales), creating information gaps that especially hurt Implicit and Visual questions. Table-as-Image and Interleaved perform similarly, with Table-as-Image slightly lower, suggesting challenges with dense or multi-image reasoning. Entity Replaced performs best among methods but still falls well below human performance, as it reduces visual ambiguity by converting images into explicit text-grounded entities, aligning better with LLM strengths and improving visual reasoning and entity disambiguation. Overall, results point to architectural limitations rather than training issues, with captioning underperforming due to loss of task-relevant detail.

\noindent
\textbf{Effect of LLMs:}
Mixtral performs best in text-only scenarios with a low unknown rate (12\%), showing that conservative uncertainty handling supports reasoning without visual input. GPT-4o Mini exhibits higher uncertainty (45\%) but excels in Entity Replaced scenarios, reflecting strength in symbolic reasoning when visual barriers are removed. Variations in “unknown” responses stem from differences in confidence calibration, risk tolerance, instruction-following, and training biases (Table~\ref{tab:unknown}). The vision-language alignment paradox appears in Qwen, which achieves the highest SS scores on Interleaved tasks but low EM, indicating strong content extraction but poor formatting. The resulting 18-point gap demonstrates that even advanced vision-language models can struggle with precise answer generation.

Additional detailed breakdowns of these results are provided in Appendix~\ref{sec:additional_results}.

\begin{table}[H]
\centering
\setlength{\aboverulesep}{0pt}
\setlength{\belowrulesep}{0pt}
\small
\setlength{\tabcolsep}{12pt}
\begin{tabular}{lc}
\toprule
\textbf{Model} & \textbf{Unknown \%} \\
\midrule
GPT-4o Mini       & 45.96\% \\
Llama 3-8B        & 41.54\% \\
Gemini 2.0 Flash  & 37.99\% \\
Gemini 1.5 Flash  & 36.23\% \\
Mixtral-8x7B      & 12.13\% \\
\bottomrule
\end{tabular}
\caption{\small Percentage of ``Unknown'' answers per model for Missing Image Baseline}
\label{tab:unknown}
\end{table}

\noindent
\textbf{Across Question Types:}
Answer-Mention questions perform best because they include direct lexical or visual references, letting models rely on simple cue matching without deeper reasoning. Explicit questions follow closely, as they still contain clear cues but require light compositional reasoning across short contexts. The narrow gap indicates that models handle shallow integration well when cues are explicit.

Implicit questions drop sharply because they demand multi-step inference and entity tracking without direct mentions, revealing weaknesses in maintaining contextual links once anchors are removed. Visual questions remain weak, as they rely on perceptual grounding rather than reasoning. While visual cues help stabilize attention, models still struggle to interpret visual details in context, making these tasks harder than those with explicit textual references.

\noindent
\textbf{Across Reasoning Types:}
Fact Verification achieves the highest performance, since these tasks resemble real-world factual checks with explicit cues. Mathematical reasoning performs the worst, highlighting persistent weaknesses in numerical logic. Even with visual context, models struggle to maintain consistency across calculation steps.

Extrema tasks perform slightly better, suggesting partial understanding of comparative or ordering relationships. Models can recognize relative patterns but often fail to generalize beyond simple comparisons. Vision-Based reasoning remains modest, as models struggle to link perceptual details with textual context. Visual cues offer some grounding, but current architectures lack the fine-grained spatial and semantic integration required for robust multimodal reasoning.

\noindent\textbf{Across Table Types:}
Single-chart performance is highest, as real-world charts are designed for quick facts with consistent legends and scales, making them easy for humans and MLLMs to interpret. Performance declines with multiple charts or chart-plus-entity setups, since the model must align scales and legends and combine information across visuals, requiring more reasoning and structural understanding.

The gap between single and multiple entity inputs is small. MLLMs, trained on many images with diverse entities, can recognize and label them easily. The main challenge lies in linking visual and textual information in structured layouts, not the number of entities.

\noindent\textbf{Across Answer Types:}
Image Location and Multiple Types perform poorly, as models struggle to integrate spatial, numeric, and entity data, which requires cross-modal reasoning that current architectures handle only partially.

Entity-based answers perform best, with Single and Multiple Entity types showing strong results. Entities are frequent in training data and easy to recognize, so models can rely on surface-level visual and lexical cues. Performance is similar for single and multiple entities since identification remains straightforward even with multiple targets.

Number-based answers are intermediate. Single Number tasks are moderately challenging, showing partial success in numeric retrieval. Multiple-number tasks perform better than expected because they do not require symbolic computation, highlighting that models struggle with mathematical reasoning but can retrieve numeric information without operations.

\begin{table*}[!htbp]
\centering
\small\setlength{\aboverulesep}{0pt}
\setlength{\belowrulesep}{0pt}
\setlength{\tabcolsep}{3pt}
\begin{tabular}{@{}l|l|ccc|ccc|ccc|ccc@{}}
\toprule
\textbf{Baseline} & \textbf{Model} & 
\multicolumn{3}{c|}{\textbf{Answer Mention}} & 
\multicolumn{3}{c|}{\textbf{Explicit}} & 
\multicolumn{3}{c|}{\textbf{Implicit}} & 
\multicolumn{3}{c}{\textbf{Visual-Based}} \\
\midrule
& & E.M & SS & F1 & E.M & SS & F1 & E.M & SS & F1 & E.M & SS & F1 \\
\midrule
Table as an Image & GPT-4o-mini & 34.8 & 38.7 & 0.32 & 44.8 & 46.9 & 0.34 & 40.7 & 42.3 & 0.32 & 41.0 & 45.2 & 0.36 \\
Interleaved & GPT-4o-mini & 41.6 & 45.9 & 0.35 & 46.4 & 47.8 & 0.37 & 31.4 & 33.0 & 0.26 & 32.2 & 34.7 & 0.27 \\
Interleaved & Human & 78.4 & 82.1 & 0.76 & 84.2 & 87.3 & 0.81 & 75.8 & 80.6 & 0.73 & 79.9 & 83.7 & 0.78 \\
\bottomrule
\end{tabular}
\vspace{-0.5em}
\caption{\small Best performing baselines on sample set evaluation.}
\label{baseline_performance}
\vspace{-0.5em}
\end{table*}

\vspace{-0.5em}

\paragraph{How does model architecture affect multimodal table reasoning} Model architecture strongly shapes multimodal table reasoning performance. Unified multimodal decoders (e.g., Gemini 1.5/2.0 Flash, GPT-4o Mini) integrate visual patches as native tokens within the transformer sequence, enabling stronger structural grounding and more reliable spatial-textual alignment across interleaved tables; this translates to the best overall reasoning and answer formatting, though challenges remain for visually grounded outputs such as image location. Dynamic-resolution ViTs (e.g., Qwen-2.5/3-VL) employ adaptive grids and 2D-RoPE \cite{heo2024rotary} to preserve layout fidelity and extract fine-grained table content, but exhibit notable gaps between semantic understanding and exact answer matching, pointing to weaker output calibration. Cross-modal projector architectures (Mantis, Phi-3.5 Vision) connect separate vision backbones to LLMs via learned connectors or resamplers, improving efficiency in multi-image settings but introducing information bottlenecks that limit precise reasoning and exact-match performance in complex tabular contexts.

\section{Human Evaluation and Error Analysis}\label{sec:HEEA_main}

We conducted human evaluation with two annotators on 50 tables (11\% of the dataset) and 393 questions (12\%), where participants answered using only their knowledge without external image search. Table ~\ref{baseline_performance} shows humans consistently outperformed all baseline models, highlighting both strong performance and the complexity of multimodal table reasoning. We analyze the best-performing model from the Table as Image and Interleaved baselines against human performance.

\paragraph{Types of Errors}

We categorize errors into several types. \textbf{(a) Entity Disambiguation Issues} occur when an image is misidentified, leading to incorrect interpretation of its content and conclusions. \textbf{(b) Entity Identification Issues} refer to cases where the image is not recognized at all, preventing meaningful analysis. \textbf{(c) Reasoning Errors} arise when the image is correctly identified but the reasoning process is flawed, resulting in incorrect answers. \textbf{(d) Identification of Visual Attributes} involves missing or misinterpreting key visual details such as shapes, colors, or patterns. \textbf{(e) Structural Errors} occur due to misunderstanding tabular or structured data, including rows, columns, or hierarchical relationships. \textbf{(f) Mathematical Errors} include mistakes in reading or computing numerical values, such as counting or calculations. \textbf{(g) Partial Answers} occur when responses are incomplete and miss essential details. \textbf{(h) Extra Information or Hallucination} involves adding incorrect or unsupported details not present in the image, often due to prior assumptions.

\begin{figure}[t]
    \centering
    \includegraphics[width=1.0\linewidth]{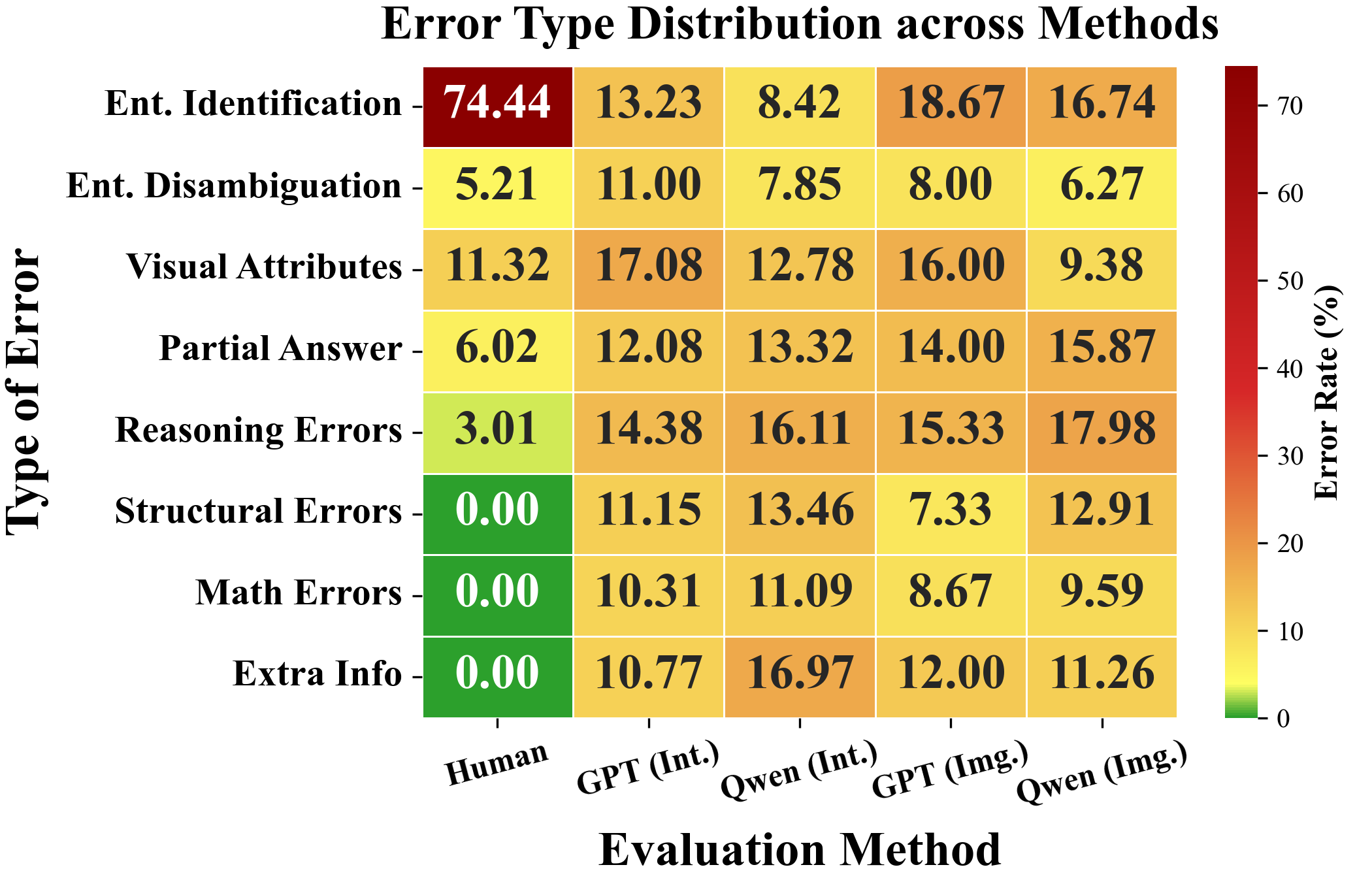}
    \caption{\small Error type distribution (in \%) across evaluation methods. Int.=Interleaved baseline, Img.=Table as Image baseline. Numbers represent percentages; column sums equal 100. Models evaluated: GPT-4o-mini and Qwen2.5-VL.}
    \label{fig:error_dist}
\end{figure}

The analysis reveals key patterns in human performance. \textbf{Entity Identification Issues} are most common, indicating difficulty in recognizing entities, especially without domain knowledge. \textbf{Partial Answers} and \textbf{Entity Disambiguation Issues} further highlight challenges in capturing complete information and distinguishing similar entities, underscoring the role of domain knowledge.

\textbf{Visual Attribute Identification Errors} show that humans can miss fine details, while \textbf{Reasoning Errors} remain low, reflecting generally strong logical processes (Figure~\ref{fig:error_dist}). Notably, there are no \textbf{Mathematical Errors}, \textbf{Structural Errors}, or \textbf{Hallucinations}, demonstrating robust numerical and structural understanding.

Overall, humans excel in reasoning and structured comprehension but struggle with entity recognition, visual detail, and completeness of answers.

Additional error-analysis details are provided in Appendix~\ref{sec:HEEA}.

\vspace{-0.3em}

\subsection{Our Key Findings: }
Our analysis reveals several specific limitations in current MLLMs. \textbf{1) Table and structured data understanding is poor}, as models struggle to interpret chart layouts, align scales, and integrate multiple types of information, resulting in very low performance on Image Location and Multiple Types answers. \textbf{2) Explicit cue recognition is strong}, with models performing better on tasks where textual or visual references are directly present, reflecting a reliance on surface-level matching rather than deeper reasoning. \textbf{3) Vision encoders are limited}, as models fail to capture subtle spatial or perceptual details in images, constraining their ability to reason over visual content. \textbf{4) Performance declines as table complexity increases}, with Single Chart tables outperforming Multiple Chart tables and Entities \& Charts showing the lowest performance, highlighting the added reasoning challenges.

\section{Comparison with Related Works}
\vspace{-0.3em}
\label{sec:prior}
As shown in Tables~\ref{tab:multimodal_structure} and~\ref{tab:multimodal_visuals} in Section~\ref{sec:motivation}, existing datasets fall short of capturing the complexity of multimodal tables across key dimensions: \begin{inparaenum}[(1)]

\vspace{0.15em}
\item \emph{Limited structural coverage:} Text-only datasets such as TaT-QA \cite{zhu2021tat}, KET-QA \cite{hu-etal-2024-ket}, FinQA, TabMWP \cite{lu2023dynamicpromptlearningpolicy} \cite{chen-etal-2021-finqa}, and DynaQA \cite{lu2022dynamic} omit visual information entirely, while early multimodal efforts—MMCoQA \cite{li-etal-2022-mmcoqa}, MMTab \cite{zheng2024multimodal}, and InfoSeek \cite{Chen2023CanPV}—include visuals only in a limited and loosely aligned manner. In contrast, visual reasoning datasets like ChartQA \cite{masry2022chartqa}, mChartQA \cite{wei2024mchartqa}, and MMC \cite{liu-etal-2024-mmc} focus on images but lack tabular structure;

\vspace{0.15em}
\item \emph{Disjoint modalities:} Datasets such as MMQA \cite{talmor2021multimodalqa}, UniMMQA \cite{luo-etal-2023-unifying}, and SPIQA \cite{yvinec2023spiq} treat text, tables, and images as separate components rather than cohesive multimodal entities, failing to capture the natural interplay of visual and textual information in real-world documents; and

\vspace{0.15em}
\item \emph{Synthetic limitations:} MMTabQA \cite{mathur-etal-2024-knowledge} moves toward multimodal integration by extending text-based datasets (e.g., FetaQA \cite{nan-etal-2022-fetaqa}, HybridQA \cite{chen-etal-2020-hybridqa}, WikiSQL \cite{zhong2017seq2sql}, Open-WikiTable \cite{kweon-etal-2023-open}) with generated visuals, but its synthetic construction and Wikipedia-only origin limit domain diversity and fail to reproduce authentic visual and structural complexity.
\end{inparaenum}

\paragraph{\datasetName vs. MMTabQA}
Among existing datasets, MMTabQA is the closest to \datasetName in scope and modality. However, \datasetName poses substantially greater challenges, as shown by uniformly lower baseline performance in Table~\ref{tab:sma-baseline-results}. The largest drops occur in the Missing Image and Image Captioning settings, indicating that MMTabQA models could more easily infer missing entities under simpler conditions.

\noindent Even the newer Gemini 2.0 Flash lags behind Gemini 1.5 Flash on our benchmark, underscoring \datasetName’s increased difficulty.

\begin{table}[!htbp]
\vspace{-0.5em}
\centering
\scriptsize
\setlength{\aboverulesep}{0pt}
\setlength{\belowrulesep}{0pt}
\setlength{\tabcolsep}{3pt}
\begin{tabular}{@{}l|ccc@{}}
\toprule
\textbf{Baseline Setting} & \textbf{MMTabQA} & \textbf{Ours} & \textbf{$\Delta$ (\%)}\\
\midrule
Missing Image (GPT 4o vs Mistral) & 62.72 & 38.85 & \textcolor{olivegreen}{-23.87} \\
Entity Replaced (GPT 4o vs GPT 4o-mini) & 78.54 & 65.95 & \textcolor{olivegreen}{-12.59} \\
Image Captioning (Gemini 1.5 vs 2.0 Flash) & 52.33 & 33.26 & \textcolor{olivegreen}{-19.07} \\
Table as Image (GPT 4o vs GPT 4o-mini) & 58.58 & 46.71 & \textcolor{olivegreen}{-11.87} \\
Interleaved (GPT 4o vs Qwen 2.5VL) & 61.08 & 53.45 & \textcolor{olivegreen}{-7.63} \\
\bottomrule
\end{tabular}
\vspace{-0.5em}
\caption{\small Baseline comparison on substring match accuracy. \datasetName results show consistently lower performance across settings.}
\label{tab:sma-baseline-results}
\vspace{-1.5em}
\end{table}

The Table-as-Image baseline declines sharply with larger tables and multiple images. The Interleaved baseline drops less, yet Qwen 2.5 VL 7B still outperforms GPT-4o on several external benchmarks (e.g., DocVQA \cite{mathew2021docvqa}, CC-OCR \cite{yang2024cc}, MathVista \cite{lu2023mathvista}, AITZ \cite{zhang2024android}, ScreenSpot \cite{li2025screenspot}), highlighting the strength of current multimodal models. Overall, the consistent degradation confirms that \datasetName is a more challenging benchmark for multimodal reasoning.

\section{Conclusion}
\vspace{-0.3em}
In conclusion, we introduce \datasetName, a benchmark designed to evaluate multimodal reasoning over real-world tables. Our experiments reveal substantial gaps in current models, with exact match scores ranging from 21.6–48.1\% and near-zero performance on Image Location and spatial reasoning tasks. Models struggle with visual-spatial reasoning, complex table structures, and tasks requiring genuine multimodal understanding, often relying on surface-level cues. \datasetName’s diverse visual and structural coverage makes it a critical benchmark for diagnosing and emphasizing the need for better integration of textual and visual information in multimodal models.

From our analysis, we suggest future avenues in multimodal table reasoning: (a) \textbf{Visual-Spatial Architectures:} Develop specialized attention mechanisms for precise spatial localization to address poor structural reasoning. (b) \textbf{Compositional Reasoning}: Develop systems capable of understanding and integrating information to perform multi-step inference and structured reasoning, going beyond simple pattern matching. (c) \textbf{Scalable Dataset Generation:} Explore automated methods that preserve authentic visual-structural complexity while expanding beyond manual curation constraints.

\section*{Limitations}
Our primary limitation is scalability due to the need for expert-driven curation. Manual creation and verification of multimodal tables ensure high-quality, semantically coherent data but make large-scale expansion challenging compared to automated generation methods. However, automated approaches would likely overlook the nuanced visual-structural relationships that define multimodal table reasoning, which are central to \datasetName’s strength. Thus, while scalability remains a constraint, it is a deliberate trade-off to preserve interpretability and evaluative precision.

A secondary limitation concerns test set size. While 500 tables may appear limited at first glance, \datasetName is designed explicitly as an evaluation benchmark rather than a training corpus. Its scale is consistent with the test set sizes of other multimodal table QA datasets, as shown in Table~\ref{tab:multimodal_structure}. Moreover, \datasetName includes a wider variety of image types, and more complex hierarchical structures, making it a richer and more diagnostic benchmark despite its focused scope.

\section*{Ethical Statement}
As the authors of this work, we confirm that our research and publication comply with the highest ethical standards. We used AI tools to assist with the writing process. The dataset used in this study is intended solely for academic purposes and should not be used for other purposes, allowing the scientific community to verify and build upon our work. All images included in the dataset are sourced from the public domain, have undergone filtering to remove sensitive content, and are free to use; even face images are publicly available and cleared for research purposes. The claims presented in this paper are consistent with the results of our experiments. To ensure reproducibility, we have provided detailed information on the prompting methods, models, and annotations used. All three reviewers involved in the inter-annotator agreement process participated voluntarily and provided informed consent.

\section*{Acknowledgement}

We gratefully acknowledge the Complex Data Analysis and Reasoning Lab and the Cognition \& Intelligence Lab at the School of Augmented Intelligence, Arizona State University, for providing computational resources and institutional support. We also thank the reviewers and external annotators for their valuable contributions.

\bibliographystyle{acl_natbib}
\bibliography{custom}
\appendix

\section{Additional Dataset and Question Types Examples} \label{sec:examples}
This section provides representative examples from each of the eight table structure types identified in Table~\ref{tab:combined_types}, demonstrating the diversity and complexity of multimodal reasoning challenges across different structural configurations.

\subsection{Single Entity}

\begin{figure}[H]
    \centering
    \includegraphics[width=1.0\linewidth]{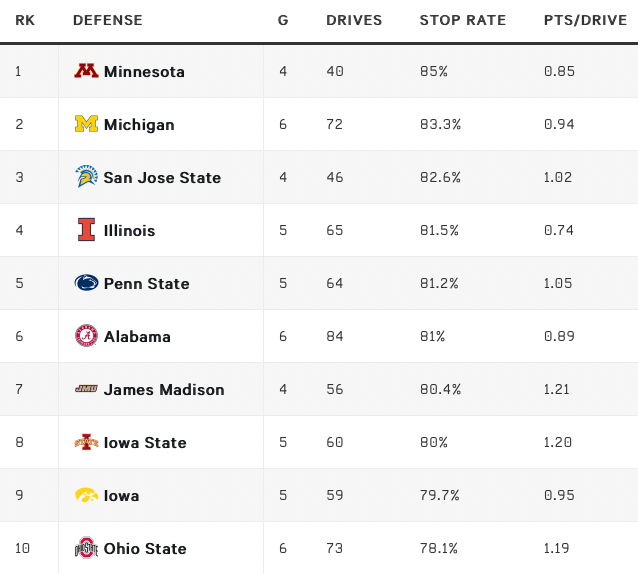}
    \caption{\small An example of single entity multimodal table in the sports domain with only one type of entity (Image of Logo+Text).}
    \label{fig:single_ent}
\end{figure}

Figure \ref{fig:single_ent} is a collegiate football defensive performance table that ranks the top ten teams based on their Stop Rate, the percentage of opponent drives that end without a score. The table includes five columns detailing performance metrics.

\noindent\textbf{Example Questions:}
\begin{itemize}[itemsep=0.1em,topsep=0.1em]
    \item[\textbf{Q:}] What is the Stop Rate for the Fighting Illini? \\
    \textbf{A:} 81.5\% \quad \textbf{Type:} Explicit.
    \vspace{-0.5em}
    \item[\textbf{Q:}] What is the difference in DRIVES faced between the highest-ranked team that played 4 Games and the highest-ranked team that played 6 Games? \\
    \textbf{A:} 32 \quad \textbf{Type:} Implicit.
    \vspace{-0.5em}
    \item[\textbf{Q:}] Which team has an allowed PTS/DRIVE that is exactly 1.21? \\
    \textbf{A:} James Madison \quad \textbf{Type:} Answer-Mention.
    \vspace{-0.5em}
    \item[\textbf{Q:}] How many teams have at least one letter from their initials in their logos? \\
    \textbf{A:} 7 \quad \textbf{Type:} Visual.
\end{itemize}

\subsection{Multiple Entity}

\begin{figure}[H]
    \centering
    \includegraphics[width=0.8\linewidth, height=0.40\textheight]{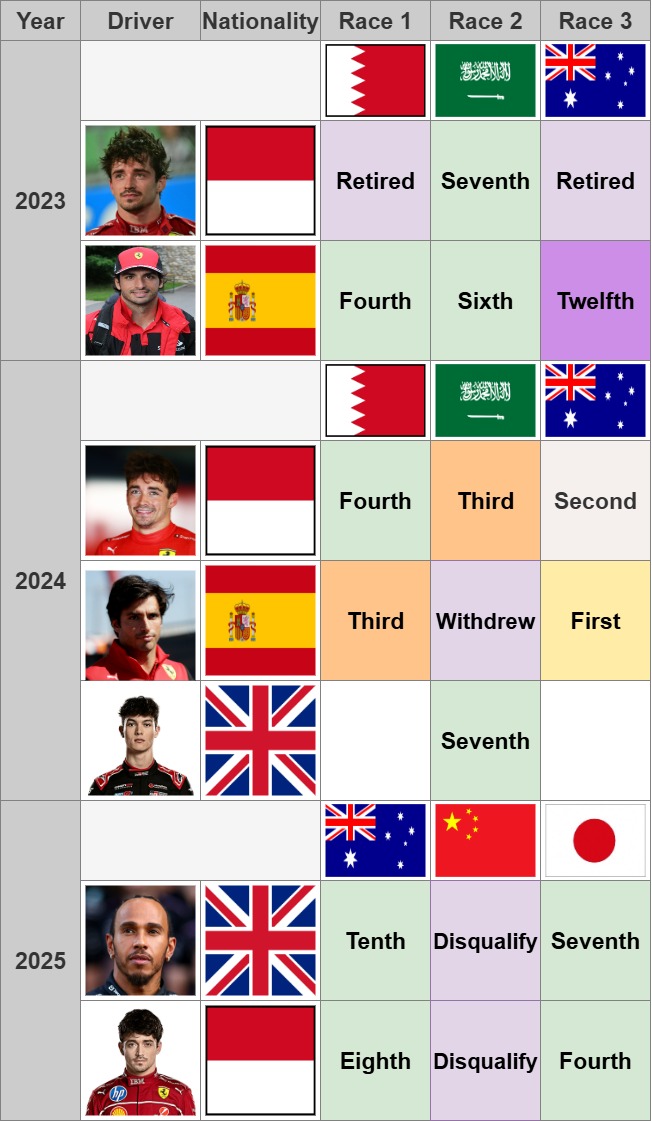}
    \caption{\small An example of multiple entity multimodal table in the motorsports domain with 2 types of entities (Image of flags + humans).}
    \label{fig:mult_ent}
\end{figure}

Figure~\ref{fig:mult_ent} is a Formula 1 Race Results Summary Table detailing the performance of multiple drivers across three consecutive racing seasons. The table is organized by Year, and each year begins with a header row displaying the Race Location Flags for the first three races of that season. The subsequent rows for each year list the Driver, their Nationality, and their finishing position in the three respective races.

\vspace{-1.0em}

\noindent\textbf{Example Questions:}
\begin{itemize}[itemsep=0.1em,topsep=0.1em]
    \item[\textbf{Q:}] How many British drivers finished in the top 5 in Australia? \\
    \textbf{A:} $0$ \quad \textbf{Type:} Explicit.
    \vspace{-0.5em}
    \item[\textbf{Q:}] What was the driver's position at the same venue one year after he finished 12th? \\
    \textbf{A:} First \quad \textbf{Type:} Implicit.
    \vspace{-0.5em}
    \item[\textbf{Q:}] Which driver replaced the driver whose national flag contains yellow, for one race? \\
    \textbf{A:} Oliver Bearman \quad \textbf{Type:} Answer-Mention.
    \vspace{-0.5em}
    \item[\textbf{Q:}] How many stars appear in this table? \\
    \textbf{A:} $23$ \quad \textbf{Type:} Visual.
\end{itemize}

\subsection{Single Chart}

\begin{figure}[H]
    \centering
    \includegraphics[width=1.0\linewidth]{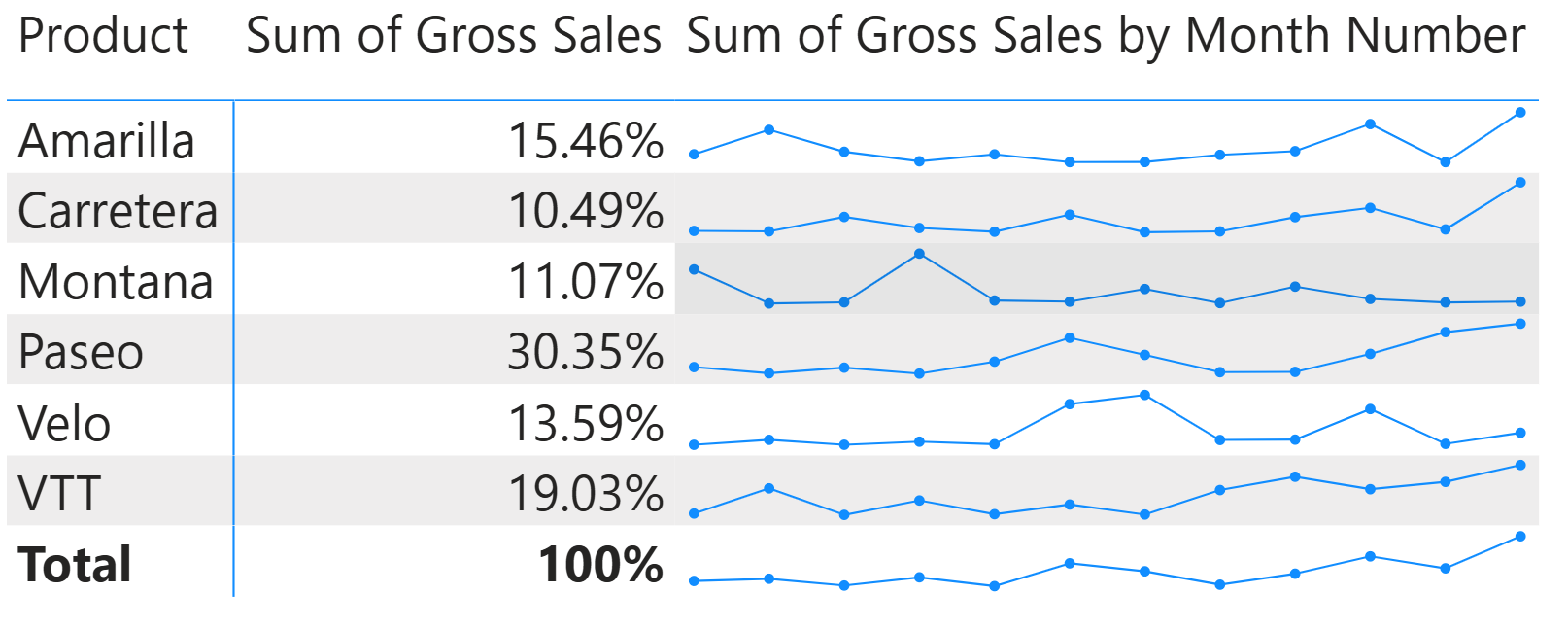}
    \caption{\small An example of a multimodal table with 1 type of chart(sparkline)}
    \label{fig:sing_chart}
\end{figure}

Figure~\ref{fig:sing_chart} is a Product Sales Performance Dashboard that displays the performance of six products (Amarilla, Carretera, Montana, Paseo, Velo, VTT) and a Total row across two metrics: Sum of Gross Sales, which shows each product's percentage contribution to total sales, and a line chart labeled Sum of Gross Sales by Month Number, which visualizes the monthly sales trend for each product.

\noindent\textbf{Example Questions:}
\begin{itemize}[itemsep=0.1em,topsep=0.1em]
    \item[\textbf{Q:}] What is the difference in Overall gross sales between the product whose sales rose the most after March vs dropped the most after March? \\
    \textbf{A:} $0.58\%$ \quad \textbf{Type:} Implicit.
    \vspace{-0.5em}
    \item[\textbf{Q:}] Which Product has the most number of drops followed by a peak? \\
    \textbf{A:} VTT \quad \textbf{Type:} Visual.
\end{itemize}

\subsection{Entities+Maps}

\begin{figure}[H]
    \centering
    \includegraphics[width=0.9\linewidth]{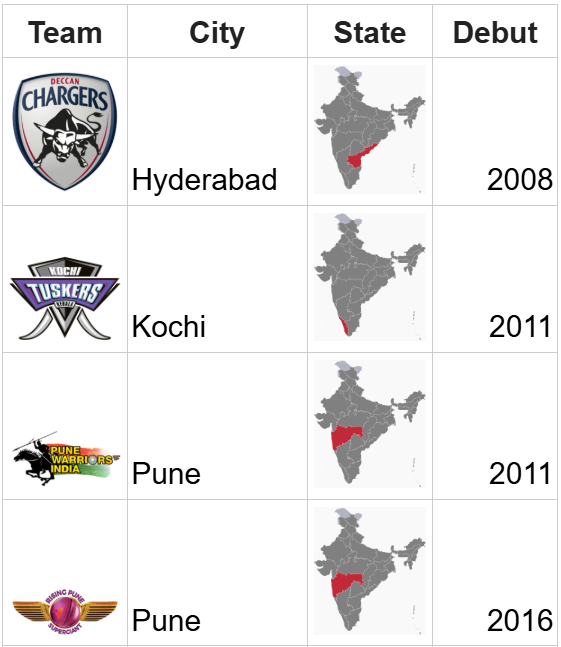}
    \caption{\small An example of a multimodal table in the sports domain with maps and entities. (Image of logos + highlighted map)}
    \label{fig:ent_map}
\end{figure}

Figure~\ref{fig:ent_map} is a table detailing four former franchises from the Indian Premier League (IPL). The table presents four columns: the team's logo and name, the home City, the State represented by a map of India with the relevant state highlighted in red, and the year the team made its league Debut.

\noindent\textbf{Example Questions:}
\begin{itemize}[itemsep=0.1em,topsep=0.1em]
    \item[\textbf{Q:}] In which year did Pune Warriors India debut? \\
    \textbf{A:} $2011$ \quad \textbf{Type:} Explicit.
    \vspace{-0.5em}
    \item[\textbf{Q:}] Which state does the team with the purple logo play in? \\
    \textbf{A:} Maharashtra \quad \textbf{Type:} Implicit.
    \vspace{-0.5em}
    \item[\textbf{Q:}] Which team from South India debuted in 2011? \\
    \textbf{A:} Kochi Tuskers Kerala \quad \textbf{Type:} Answer-Mention.
    \vspace{-0.5em}
    \item[\textbf{Q:}] How many teams have animals in their logo? \\
    \textbf{A:} $3$ \quad \textbf{Type:} Visual.
\end{itemize}

\subsection{Entities+Charts}

\begin{figure}[H]
    \centering
    \includegraphics[width=0.95\linewidth]{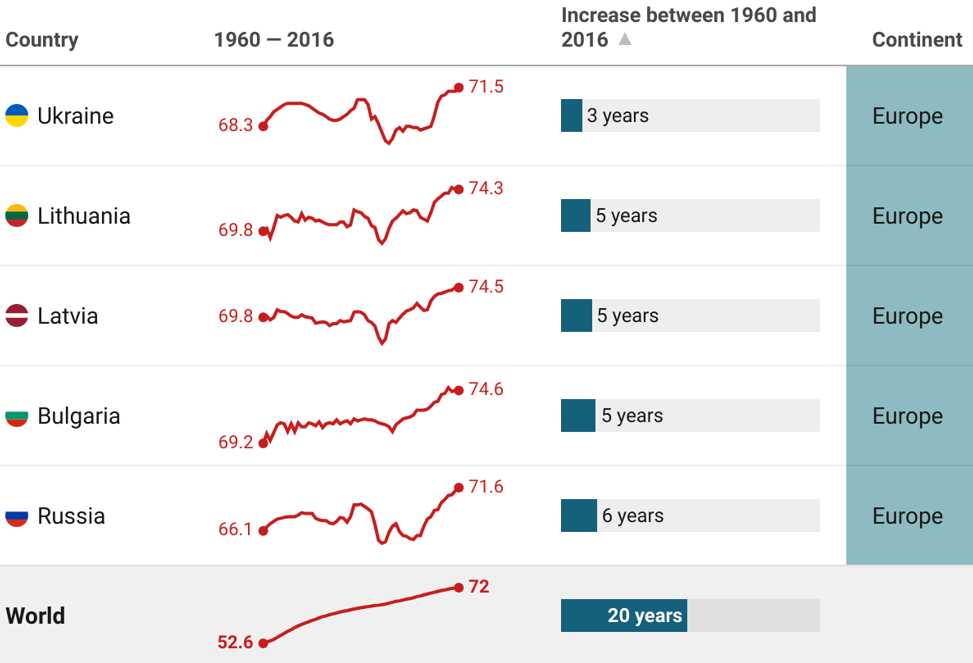}
    \caption{\small An example of a multimodal table in the Social Sciences Domain with charts and entities (Images of country flags, sparkline chart, and bar charts).}
    \label{fig:ent_chart}
\end{figure}

Figure~\ref{fig:ent_chart} is a data visualization comparing the change in life expectancy across five Eastern European countries and the global average between 1960 and 2016. The table features four columns: Country, a line chart showing the trend, the total Increase between 1960 and 2016, and the Continent.

\noindent\textbf{Example Questions:}
\begin{itemize}[itemsep=0.1em,topsep=0.1em]
    \item[\textbf{Q:}] Has the score of Ukraine ever dropped below its starting point? \\
    \textbf{A:} Yes \quad \textbf{Type:} Explicit.
    \vspace{-0.5em}
    \item[\textbf{Q:}] By how many percent did the global data increase throughout the years? \\
    \textbf{A:} $36.88\%$ \quad \textbf{Type:} Implicit.
    \vspace{-0.5em}
    \item[\textbf{Q:}] Which Country had the highest peak? \\
    \textbf{A:} Bulgaria \quad \textbf{Type:} Answer-Mention.
    \vspace{-0.5em}
    \item[\textbf{Q:}] By how many points did the country whose flag does not contain any red grow? \\
    \textbf{A:} $3.2$ \quad \textbf{Type:} Visual.
\end{itemize}

\subsection{Maps Only}

\begin{figure}[H]
    \centering
    \includegraphics[width=0.6\linewidth]{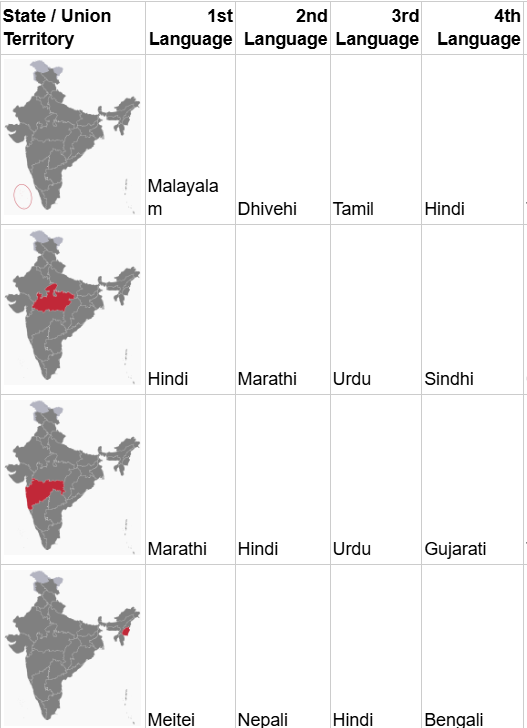}
    \caption{\small An example of a multimodal table in the Linguistics Domain with maps}
    \label{fig:maps}
\end{figure}

Figure~\ref{fig:maps} is a table that displays the four most spoken languages in four different States or Union Territories of India. The table features five columns: an image column showing a map of India with the relevant State / Union Territory highlighted in red, followed by four columns listing the 1st, 2nd, 3rd, and 4th most spoken Language in that region.

\noindent\textbf{Example Questions:}
\begin{itemize}[itemsep=0.1em,topsep=0.1em]
    \item[\textbf{Q:}] What is Lakshadweep's third language? \\
    \textbf{A:} Tamil \quad \textbf{Type:} Explicit.
    \vspace{-0.5em}
    \item[\textbf{Q:}] What is the first language of the region that borders Assam? \\
    \textbf{A:} Meitei \quad \textbf{Type:} Implicit.
    \vspace{-0.5em}
    \item[\textbf{Q:}] Name all states/union territories that speak Hindi? \\
    \textbf{A:} [Lakshadweep, Madhya Pradesh, Maharashtra, Manipur] \quad \textbf{Type:} Answer-Mention.
    \vspace{-0.5em}
    \item[\textbf{Q:}] How many states are completely landlocked? \\
    \textbf{A:} $2$ \quad \textbf{Type:} Visual.
\end{itemize}

\subsection{Program of Thought and Code Generation}

We acknowledge that Program-of-Thought (PoT) or code-generation approaches have demonstrated strong performance on tabular reasoning tasks. However, these methods typically assume that tables are in a normalized, SQL-like format with clear, flat column headers. In contrast, our dataset contains many non-standard structures, including merged cells, irregular and hierarchical headers, and other layout complexities—challenges exemplified by tables such as the multiple entity example in Figure~\ref{fig:mult_ent}, which features hierarchical organization across multiple entities and races.

Without a robust table transformation or normalization process to convert such tables into a structured, relational format, code-based methods would struggle to interpret the data correctly. Consequently, using standard PoT/code-generation approaches as a baseline would not provide a fair or meaningful comparison for our task, as the computational overhead and preprocessing complexity required to handle such diverse structural variations would make results difficult to interpret and reproduce.

\section{Inter-Annotator Guidelines}\label{sec:IAA}

To ensure dataset quality and reliability, we implemented a comprehensive multi-stage annotation process with independent review and validation. Our approach involved two phases of inter-annotator agreement analysis to assess question correctness across all 4,021 questions.

\textbf{Question Correctness Evaluation Criteria:} All annotators used a standardized 3-point scale for question correctness evaluation with detailed guidelines to ensure consistency:

\begin{itemize}[itemsep=0.1em]
    \item \textbf{Score 0 (Incorrect):} Questions that contain factual errors, ambiguous phrasing that prevents accurate answering, or answers that cannot be derived from the provided table content. This includes questions with incorrect references to table elements, logical inconsistencies, or answers that contradict visual or textual information in the table.
    
    \item \textbf{Score 1 (Partially Correct):} Questions that are generally well-formed but have minor issues such as slight ambiguity in phrasing, answers that are approximately correct but lack precision, or questions that could benefit from clearer wording. These questions are answerable but may require interpretation or have multiple plausible answers.
    
    \item \textbf{Score 2 (Fully Correct):} Questions that are clearly formulated, unambiguous, and have definitive answers that can be accurately derived from the table content. These questions demonstrate appropriate difficulty level, require genuine multimodal reasoning, and align perfectly with the provided visual and textual information.
\end{itemize}

\textbf{Table Filtering Criteria:} Annotators evaluated tables using comprehensive guidelines to ensure dataset quality and appropriateness for multimodal reasoning evaluation:

\begin{itemize}[itemsep=0.1em]
    \item \textbf{Structural Coherence Requirements:} Tables must demonstrate logical organization with clear relationships between textual and visual components. The structure should support meaningful multimodal reasoning tasks, with appropriate alignment between headers, data cells, and embedded visual elements.

\item \textbf{Content Complexity Thresholds:} Tables must contain sufficient complexity to warrant multimodal analysis, including diverse visual elements (charts, maps, images), hierarchical data organization, and content that requires integration of both visual and textual information for comprehension.

\item \textbf{Reasoning Appropriateness:} Tables should enable authentic multimodal reasoning scenarios where visual elements are essential for answering questions, rather than serving merely decorative purposes. The content must support various question types including fact verification, mathematical calculation, extrema identification, and visual-based inference.

\item \textbf{Profanity and Sensitivity Compliance:} Tables must not contain or depict any profane, explicit, hateful, or sensitive content, including personally identifiable information (PII) or imagery that violates ethical, cultural, or privacy standards. All content should be appropriate for general academic and research use.

\item \textbf{Data Source and Open-Domain Validity:} All table data and visual content should originate from credible, open-domain, and publicly accessible sources. Datasets must be free of copyright or usage restrictions that prevent open dissemination, ensuring reproducibility and transparency in multimodal research.

\end{itemize}

Cohen's Kappa ($\kappa$) was computed on independent judgments to measure inherent agreement across both evaluation criteria.

\paragraph{Phase 1 - Internal Review:} 

Table~\ref{tab:answer_agreement} reports Cohen's Kappa and Table ~\ref{tab:confusion_internal} provides the confusion matrix for those scores.

\begin{table}[htbp]
\centering
\small
\begin{tabular}{c|ccc|c}
\toprule
& \multicolumn{3}{c|}{\textbf{Annotator B}} & \\
\textbf{Annotator A} & \textbf{0} & \textbf{1} & \textbf{2} & \textbf{Total} \\
\midrule
\textbf{0} & 156 & 18 & 4 & 178 \\
\textbf{1} & 22 & 298 & 45 & 365 \\
\textbf{2} & 8 & 51 & 3419 & 3478 \\
\midrule
\textbf{Total} & 186 & 367 & 3468 & 4021 \\
\bottomrule
\end{tabular}
\caption{Confusion Matrix: Internal Expert Annotators (Phase 1)}
\label{tab:confusion_internal}
\end{table}

\paragraph{Phase 2 - External Validation (Subset):}
Table~\ref{tab:third_reviewer_agreement} reports Cohen's Kappa and Table ~\ref{tab:confusion_external} provides the confusion matrix for those scores.

\begin{table}[htbp]
\centering
\small
\begin{tabular}{c|ccc|c}
\toprule
& \multicolumn{3}{c|}{\textbf{Gold Standard}} & \\
\textbf{External Annotators} & \textbf{0} & \textbf{1} & \textbf{2} & \textbf{Total} \\
\midrule
\textbf{0} & 32 & 3 & 1 & 36 \\
\textbf{1} & 4 & 71 & 8 & 83 \\
\textbf{2} & 2 & 9 & 674 & 685 \\
\midrule
\textbf{Total} & 38 & 83 & 683 & 804 \\
\bottomrule
\end{tabular}
\caption{Confusion Matrix: External Annotators vs Gold Standard (Phase 2)}
\label{tab:confusion_external}
\end{table}

\section{Additional Result Tables}
\label{sec:additional_results}

This section presents detailed performance results across all reasoning types and answer types discussed in Section~\ref{sec:experiments}.

Tables~\ref{tab:reasoning_types_results} and ~\ref{tab:answer_types_results} provide comprehensive performance breakdowns across all reasoning types and answer formats.

\begin{table*}[h]
\vspace{-0.5em}
\scriptsize
\centering
\setlength{\aboverulesep}{0pt}
\setlength{\belowrulesep}{0pt}
\setlength{\tabcolsep}{7.5pt}
\begin{tabular}{lccc|ccc|ccc|ccc}
\toprule
\multicolumn{1}{c}{} & \multicolumn{3}{c|}{\textbf{Fact Verification}} & \multicolumn{3}{c|}{\textbf{Mathematical}} & \multicolumn{3}{c|}{\textbf{Extrema}} & \multicolumn{3}{c}{\textbf{Vision Based}} \\
\cmidrule{2-13}
Model & EM & SS & F1 & EM & SS & F1 & EM & SS & F1 & EM & SS & F1 \\
\cmidrule{2-13}
& \multicolumn{12}{c}{\textbf{Missing Image Baseline}} \\
\cmidrule{2-13}
\midrule
Gemini 1.5 Flash & 28.86 & 29.85 & 0.179 & 16.42 & 17.52 & 0.059 & 16.73 & 17.47 & 0.070 & 5.23 & 5.52 & 0.030 \\
Gemini 2.0 Flash & 27.62 & 27.61 & 0.099 & 15.96 & 16.35 & 0.049 & 15.80 & 16.20 & 0.059 & 11.00 & 11.04 & 0.032 \\
GPT-4o Mini     & 44.75 & 47.64 & 0.361 & 23.80 & 23.29 & 0.124 & 27.45 & 27.07 & 0.146 & 24.09 & 25.64 & 0.177 \\
Llama 3-8B      & 38.48 & 40.60 & 0.258 & 20.25 & 22.18 & 0.095 & 18.15 & 20.17 & 0.115 & 23.01 & 23.85 & 0.168 \\
Mixtral    & \cellcolor{yellow}47.43 & \cellcolor{yellow}52.33 & \cellcolor{yellow}0.429 & \cellcolor{yellow}24.66 & \cellcolor{yellow}30.91 & \cellcolor{yellow}0.174 & \cellcolor{yellow}27.29 & \cellcolor{yellow}32.58 & \cellcolor{yellow}0.184 & \cellcolor{yellow}29.07 & \cellcolor{yellow}33.38 & \cellcolor{yellow}0.235 \\
\cmidrule{2-13}
& \multicolumn{12}{c}{\textbf{Entity Replaced Baseline}} \\
\cmidrule{2-13}
Gemini 1.5 Flash & 68.18 & 72.65 & 0.322 & 40.10 & 42.48 & 0.210 & 57.01 & 60.21 & 0.324 & - & - & - \\
Gemini 2.0 Flash & 71.69 & 73.28 & 0.333 & \cellcolor{yellow}42.11 & 41.20 & 0.191 & \cellcolor{yellow}64.80 & \cellcolor{yellow}65.86 & 0.347 & - & - & - \\
GPT-4o Mini     & 79.45 & 75.11 & 0.628 & 39.00 & \cellcolor{yellow}43.92 & \cellcolor{yellow}0.255 & 54.06 & 57.00 & \cellcolor{yellow}0.357 & - & - & - \\
Llama 3-8B      & 70.45 & 74.84 & 0.586 & 32.15 & 37.40 & 0.205 & 41.02 & 44.32 & 0.270 & - & - & - \\
Mixtral    & \cellcolor{yellow}80.17 & \cellcolor{yellow}83.34 & \cellcolor{yellow}0.643 & 34.05 & 39.97 & 0.237 & 44.16 & 49.18 & 0.317 & - & - & - \\
\cmidrule{2-13}
& \multicolumn{12}{c}{\textbf{Image Captioning Baseline}} \\
\cmidrule{2-13}
Gemini 1.5 Flash & 47.90 & 49.29 & \cellcolor{yellow}0.369 & 13.68 & 15.38 & 0.074 & 25.13 & 27.47 & 0.165 & 22.72 & 23.76 & 0.186 \\
Gemini 2.0 Flash & \cellcolor{yellow}48.14 & \cellcolor{yellow}51.87 & 0.358 & \cellcolor{yellow}19.60 & \cellcolor{yellow}21.43 & \cellcolor{yellow}0.092 & \cellcolor{yellow}28.48 & \cellcolor{yellow}33.31 & \cellcolor{yellow}0.179 & \cellcolor{yellow}29.15 & \cellcolor{yellow}30.22 & \cellcolor{yellow}0.235 \\
\cmidrule{2-13}
& \multicolumn{12}{c}{\textbf{Table as an Image Baseline}} \\
\cmidrule{2-13}
Gemini 1.5 Flash & 47.17 & 45.77 & 0.207 & 23.42 & 25.33 & 0.095 & 21.30 & 23.01 & 0.113 & 30.40 & 32.38 & 0.094 \\
Gemini 2.0 Flash & 45.75 & 46.04 & 0.254 & 29.57 & 32.16 & 0.174 & 29.92 & 30.27 & 0.179 & 29.11 & 30.72 & 0.163 \\
GPT-4o Mini      & \cellcolor{yellow}63.34 & \cellcolor{yellow}64.06 & \cellcolor{yellow}0.490 & \cellcolor{yellow}34.00 & 36.52 & \cellcolor{yellow}0.214 & \cellcolor{yellow}41.62 & \cellcolor{yellow}43.86 & \cellcolor{yellow}0.275 & \cellcolor{yellow}38.50 & \cellcolor{yellow}41.45 & \cellcolor{yellow}0.306 \\
InternVL 2.5-8B  & 19.94 & 46.19 & 0.222 & 15.80 & \cellcolor{yellow}38.13 & 0.154 & 15.94 & 33.66 & 0.148 & 11.76 & 38.97 & 0.141 \\
Mantis-8B-Idefics2 & 23.28 & 27.00 & 0.155 & 19.63 & 22.24 & 0.084 & 15.54 & 16.20 & 0.071 & 19.07 & 22.51 & 0.118 \\
Phi-3.5-vision-instruct & 32.16 & 35.79 & 0.238 & 13.87 & 16.18 & 0.052 & 13.73 & 15.01 & 0.063 & 15.24 & 17.70 & 0.075 \\
Qwen-2.5-VL & 53.12 & 55.19 & 0.265 & 21.45 & 25.08 & 0.113 & 27.90 & 31.39 & 0.158 & 22.54 & 24.40 & 0.102 \\
Qwen-3-VL        & 58.20 & 60.50 & 0.380 & 28.50 & 31.20 & 0.165 & 35.80 & 38.40 & 0.220 & 31.80 & 34.20 & 0.215 \\
Table-llava-1.5-7b-hf & 24.57 & 27.06 & 0.133 & 14.57 & 15.51 & 0.038 & 8.42 & 9.54 & 0.031 & 9.88 & 10.93 & 0.049 \\
\cmidrule{2-13}
& \multicolumn{12}{c}{\textbf{Interleaved Baseline}} \\
\cmidrule{2-13}
Gemini 1.5 Flash & 41.37 & 41.14 & 0.283 & 17.96 & 18.00 & 0.079 & 27.78 & 27.21 & 0.155 & 26.81 & 27.28 & 0.213 \\
Gemini 2.0 Flash & 43.66 & 48.04 & 0.332 & 20.08 & 21.56 & 0.092 & 30.18 & 32.88 & 0.153 & 31.51 & 32.62 & 0.232 \\
GPT-4o Mini     & \cellcolor{yellow}61.40 & 61.91 & \cellcolor{yellow}0.479 & \cellcolor{yellow}28.84 & 29.66 & \cellcolor{yellow}0.174 & \cellcolor{yellow}38.78 & 40.97 & \cellcolor{yellow}0.265 & \cellcolor{yellow}39.69 & 41.87 & \cellcolor{yellow}0.316 \\
Mantis-8B-Idefics2 & 29.13 & 32.50 & 0.190 & 20.01 & 21.94 & 0.082 & 18.83 & 20.40 & 0.097 & 27.06 & 26.81 & 0.144 \\
Phi-3.5-vision-instruct & 25.67 & 28.31 & 0.168 & 21.98 & 24.72 & 0.107 & 16.39 & 17.39 & 0.093 & 17.33 & 19.14 & 0.106 \\
Qwen-2.5-VL & 43.03 & \cellcolor{yellow}62.76 & 0.381 & 15.77 & \cellcolor{yellow}48.33 & 0.106 & 24.90 & \cellcolor{yellow}52.69 & 0.198 & 25.99 & \cellcolor{yellow}52.47 & 0.225 \\
Qwen-3-VL & 38.50 & 61.80 & 0.310 & 14.20 & 26.50 & 0.085 & 22.40 & 34.60 & 0.155 & 24.30 & 36.80 & 0.190 \\
\bottomrule
\end{tabular}
\vspace{-0.5em}
\caption{Performance analysis across reasoning types. EM: Exact Match, SS: Substring Match, F1: F1 Score}
\label{tab:reasoning_types_results}
\vspace{-0.5em}
\end{table*}

\begin{table*}[h]
\vspace{-0.5em}
\scriptsize
\centering
\setlength{\aboverulesep}{0pt}
\setlength{\belowrulesep}{0pt}
\setlength{\tabcolsep}{8.5pt}
\begin{tabular}{lcc|cc|cc|cc|cc|cc}
\toprule
\multicolumn{1}{c}{} & \multicolumn{2}{c|}{\textbf{SE}} & \multicolumn{2}{c|}{\textbf{ME}} & \multicolumn{2}{c|}{\textbf{SN}} & \multicolumn{2}{c|}{\textbf{MN}} & \multicolumn{2}{c|}{\textbf{IL}} & \multicolumn{2}{c}{\textbf{MT}} \\
\cmidrule{2-13}
Model & EM & F1 & EM & F1 & EM & F1 & EM & F1 & EM & F1 & EM & F1 \\
\cmidrule{2-13}
& \multicolumn{12}{c}{\textbf{Missing Image Baseline}} \\
\cmidrule{2-13}
Gemini 1.5 Flash 
 & 18.9   & 0.071 
 & 21.1   & 0.081 
 & 13.3   & 0.037 
 & 20.4   & 0.084 
 & \cellcolor{yellow}15.4 & 0.028 
 & 22.1   & 0.112 \\
Gemini 2.0 Flash 
 & 19.0   & 0.084 
 & 16.9   & 0.094 
 & 15.8   & 0.037 
 & 24.8   & 0.090 
 & 2.6    & 0.014 
 & 25.9   & 0.067 \\
GPT-4o Mini 
 & 32.0   & 0.230 
 & 32.0   & 0.201 
 & 27.4   & 0.188 
 & 33.0   & 0.224 
 & 10.6   & \cellcolor{yellow}0.073 
 & 37.3   & 0.240 \\
Llama 3-8B 
 & 28.1   & 0.188 
 & 29.7   & 0.182 
 & 21.3   & 0.137 
 & 29.1   & 0.145 
 & 0.0    & 0.000 
 & 25.4   & 0.160 \\
Mixtral 
 & \cellcolor{yellow}35.3 & \cellcolor{yellow}0.273 
 & \cellcolor{yellow}33.3 & \cellcolor{yellow}0.247 
 & \cellcolor{yellow}28.4 & \cellcolor{yellow}0.209 
 & \cellcolor{yellow}36.7 & \cellcolor{yellow}0.233 
 & 3.4    & 0.025 
 & \cellcolor{yellow}38.0 & \cellcolor{yellow}0.316 \\
\cmidrule{2-13}
& \multicolumn{12}{c}{\textbf{Entity Replaced Baseline}} \\
\cmidrule{2-13}
Gemini 1.5 Flash 
 & 37.8   & 0.114 
 & 57.4   & 0.385 
 & 7.2    & \cellcolor{yellow}0.615 
 & 25.7   & 0.193 
 & 8.0    & \cellcolor{yellow}0.300 
 & \cellcolor{yellow}36.7 & 0.308 \\
Gemini 2.0 Flash 
 & 41.7   & 0.211 
 & 59.0   & 0.315 
 & 15.8   & 0.558 
 & 30.8   & 0.248 
 & 11.0   & 0.167 
 & 16.7   & 0.257 \\
GPT-4o Mini 
 & \cellcolor{yellow}48.1 & 0.290 
 & \cellcolor{yellow}66.9 & \cellcolor{yellow}0.481 
 & 22.8   & 0.598 
 & \cellcolor{yellow}36.8 & 0.334 
 & 13.4   & 0.227 
 & 24.1   & \cellcolor{yellow}0.368 \\
Llama 3-8B 
 & 39.8   & 0.210 
 & 57.5   & 0.412 
 & 16.2   & 0.504 
 & 31.7   & 0.308 
 & 13.8   & 0.106 
 & 14.2   & 0.317 \\
Mixtral 
 & 41.3   & \cellcolor{yellow}0.293 
 & 56.6   & 0.446 
 & \cellcolor{yellow}29.7 & 0.593 
 & 35.0   & \cellcolor{yellow}0.348 
 & \cellcolor{yellow}14.8 & 0.205 
 & 27.7   & 0.350 \\
\cmidrule{2-13}
& \multicolumn{12}{c}{\textbf{Image Captioning Baseline}} \\
\cmidrule{2-13}
Gemini 1.5 Flash 
 & 28.3   & 0.217 
 & 22.7   & 0.155 
 & 25.3   & \cellcolor{yellow}0.188 
 & 12.9   & 0.051 
 & \cellcolor{yellow}5.8 & 0.058 
 & 14.7   & 0.056 \\
Gemini 2.0 Flash 
 & \cellcolor{yellow}33.3 & \cellcolor{yellow}0.243 
 & \cellcolor{yellow}34.5 & \cellcolor{yellow}0.230 
 & \cellcolor{yellow}25.8 & 0.183 
 & \cellcolor{yellow}23.2 & \cellcolor{yellow}0.109 
 & 4.0    & \cellcolor{yellow}0.029 
 & \cellcolor{yellow}29.6 & \cellcolor{yellow}0.131 \\
\cmidrule{2-13}
& \multicolumn{12}{c}{\textbf{Table As Image Baseline}} \\
\cmidrule{2-13}
Gemini 1.5 Flash 
 & 38.0   & 0.223 
 & 41.6   & 0.274 
 & 21.6   & 0.078 
 & 46.6   & 0.255 
 & 23.2   & 0.113 
 & 34.3   & 0.199 \\
Gemini 2.0 Flash 
 & 30.3   & 0.147 
 & 35.6   & 0.168 
 & 19.7   & 0.072 
 & 38.6   & 0.174 
 & \cellcolor{yellow}26.9 & 0.119 
 & 32.5   & 0.149 \\
GPT-4o Mini 
 & \cellcolor{yellow}47.0 & \cellcolor{yellow}0.355 
 & \cellcolor{yellow}46.5 & \cellcolor{yellow}0.305 
 & \cellcolor{yellow}33.4 & \cellcolor{yellow}0.228 
 & \cellcolor{yellow}50.1 & \cellcolor{yellow}0.318 
 & 26.7   & \cellcolor{yellow}0.214 
 & \cellcolor{yellow}42.5 & \cellcolor{yellow}0.304 \\
Intern-VL-2.5 
 & 17.7   & 0.187 
 & 18.9   & 0.189 
 & 12.1   & 0.125 
 & 27.6   & 0.292 
 & 8.6    & 0.080 
 & 23.0   & 0.227 \\
Mantis-8B-Idefics2 
 & 22.6   & 0.140 
 & 23.8   & 0.106 
 & 10.7   & 0.040 
 & 24.6   & 0.088 
 & 25.6   & 0.122 
 & 22.7   & 0.083 \\
Phi-3.5 
 & 18.2   & 0.089 
 & 21.2   & 0.111 
 & 12.5   & 0.038 
 & 17.6   & 0.023 
 & 0.1    & 0.002 
 & 14.0   & 0.056 \\
Qwen-2.5-VL 
 & 29.7   & 0.160 
 & 32.1   & 0.190 
 & 15.3   & 0.059 
 & 29.8   & 0.161 
 & 14.1   & 0.067 
 & 29.2   & 0.181 \\
Qwen-3-VL 
 & 38.5   & 0.265 
 & 39.8   & 0.250 
 & 24.0   & 0.145 
 & 40.2   & 0.245 
 & 21.0   & 0.145 
 & 36.5   & 0.245 \\
Table\_LLaVA 
 & 12.9   & 0.078 
 & 14.0   & 0.050 
 & 10.8   & 0.030 
 & 21.9   & 0.065 
 & 0.0    & 0.000 
 & 16.6   & 0.050 \\
\cmidrule{2-13}
& \multicolumn{12}{c}{\textbf{Interleaved Baseline}} \\
\cmidrule{2-13}
Gemini 1.5 Flash 
 & 29.2   & 0.205 
 & 29.0   & 0.177 
 & 25.2   & 0.175 
 & 25.2   & 0.120 
 & 4.5    & 0.034 
 & 26.6   & 0.131 \\
Gemini 2.0 Flash 
 & 33.2   & 0.223 
 & 34.6   & 0.210 
 & 27.9   & 0.195 
 & 29.4   & 0.156 
 & 26.6   & 0.131 
 & \cellcolor{yellow}27.3 & 0.103 \\
GPT-4o Mini 
 & \cellcolor{yellow}48.1 & \cellcolor{yellow}0.372 
 & \cellcolor{yellow}47.7 & \cellcolor{yellow}0.301 
 & \cellcolor{yellow}30.9 & \cellcolor{yellow}0.223 
 & \cellcolor{yellow}40.0 & \cellcolor{yellow}0.252 
 & 21.6   & 0.127 
 & 24.4   & 0.184 \\
Mantis-8B-Idefics2 
 & 26.0   & 0.175 
 & 29.2   & 0.144 
 & 15.6   & 0.064 
 & 15.8   & 0.061 
 & \cellcolor{yellow}27.0 & \cellcolor{yellow}0.139 
 & 21.7   & 0.066 \\
Phi-3.5 
 & 21.4   & 0.136 
 & 27.7   & 0.144 
 & 13.7   & 0.060 
 & 28.8   & 0.131 
 & 13.7   & 0.071 
 & 27.0 & \cellcolor{yellow}0.185 \\
Qwen-2.5-VL 
 & 29.4   & 0.261 
 & 21.5   & 0.177 
 & 23.5   & 0.185 
 & 22.0   & 0.158 
 & 5.7    & 0.060 
 & 20.2   & 0.165 \\
Qwen-3-VL 
 & 39.0   & 0.320 
 & 35.5   & 0.240 
 & 27.0   & 0.205 
 & 31.0   & 0.210 
 & 19.5   & 0.105 
 & 23.8   & 0.175 \\
\bottomrule
\end{tabular}
\vspace{-0.5em}
\caption{Performance across answer types. SE: Single Entity, ME: Multiple Entity, SN: Single Number, MN: Multiple Number, IL: Image Location, MT: Multiple Types.}
\label{tab:answer_types_results}
\vspace{-0.5em}
\end{table*}

\section{Human Evaluation and Error Analysis}\label{sec:HEEA}

While responses may contain multiple errors, we adopt a hierarchical approach to identify the \textit{root cause} of failure, avoiding double-counting of derivative errors. When multiple issues occur, each instance is attributed only to the primary cause. For example, if a visual attribute is missed and leads to an incomplete answer, the error is classified as \textit{Identification of Visual Attributes}, rather than a reasoning or partial answer error.

\subsection{Performance and Analysis of Errors in Interleaved baseline GPT 4o-mini}

GPT-4o-mini exhibits significant errors across multiple categories, with each exceeding the 10\% threshold. Reasoning Errors are the most prominent, as the model struggles with multi-step deductions and logical coherence - an ongoing challenge for MLLMs. Visual Attribute Identification is another major limitation, with failures in extracting key image features, underscoring the need for better vision encoders and fine-tuning.

Entity Identification and Disambiguation errors occur at similar rates, as the model misidentifies or fails to recognize entities, likely due to insufficient training data and over-reliance on context. Structural Errors show difficulty in interpreting complex tabular data, including hierarchical structures and nested tables, while Hallucination Errors highlight the model’s tendency to generate irrelevant information. Mathematical Errors reveal persistent struggles with quantitative reasoning.

These issues demonstrate that while GPT-4o-mini has some multimodal reasoning capability, it lacks depth in entity recognition, visual interpretation, and structured data comprehension.

\subsection{Performance and Analysis of Errors in Table as an Image baseline GPT 4o-mini}
GPT4o-mini makes many errors across all categories. From figure \ref{fig:error_dist}, in table as image baseline, we notice that GPT4o-mini had most errors in Entity Identification. This is likely due to insufficient training data to much reliance on context.

It also struggles with Identification of visual attributes and reasoning errors in similar range as Interleaved baseline. We see similar error rates among Mathematical, Hallucinations, Partial and Entity Disambiguation error types showing that GPT4o-mini struggles with these issues irrespective of input table format.

Interestingly, We observe that structural Errors decreased by \textasciitilde4\% in table as image baseline. This may be due to the fact that looking at entire table at once through image helped understand the structure better. So, passing tables as image does help GPT4o-mini to understand the structure of tables better.

From above both analysis we can say GPT4o-mini has scope of improvement in identifying visual attributes, on entity identification and reasoning errors.

\subsection{Performance and Analysis of Errors in Interleaved baseline Qwen2.5-VL}
The error distribution in Qwen2.5-VL suggests a model that demonstrates broad multimodal competence but lacks consistent grounding in visual inputs. The high incidence of hallucination-related errors (16.97\%) implies that the model frequently relies on pretrained textual priors rather than accurately integrating perceptual evidence. This points to limitations in the vision-language alignment mechanisms, where generated responses may reflect plausible but unsubstantiated inferences.

Reasoning errors (16.11\%) and structural errors (13.46\%) indicate significant challenges in multi-step logical processing and the interpretation of complex visual structures, such as nested tables or hierarchical layouts. These errors suggest that while Qwen2.5-VL is able to extract surface-level patterns, it struggles with deeper, context-sensitive reasoning tasks - an area where multimodal systems continue to underperform.

Although errors in entity identification (8.42\%) and disambiguation (7.85\%) are comparatively lower, they reflect persistent difficulties in maintaining contextual precision, particularly in visually dense or ambiguous scenarios. Overall, the model’s performance underscores the need for improved grounding strategies, enhanced visual encoding, and tighter integration of reasoning capabilities to support more reliable and context-aware multimodal understanding.

\subsection{Performance and Analysis of Errors in Table as an Image baseline Qwen2.5-VL}

Qwen2.5-VL’s performance on table-as-image tasks highlights persistent challenges in structured data interpretation. The most prominent error category is Reasoning Errors (17.98\%), suggesting the model struggles to perform logical operations over tabular data, particularly when spatial layout or implicit relationships between rows and columns are involved. This reflects a broader difficulty in translating visual table structures into coherent semantic representations.

Partial Answers (15.87\%) and Entity Identification Issues (16.74\%) further indicate that Qwen2.5-VL often fails to capture complete and precise information from table-based inputs. These errors likely stem from limitations in visual attention mechanisms or token alignment between image features and textual output. Similarly, the model’s relatively high rate of Visual Attribute Identification errors (9.38\%) and Structural Errors (12.91\%) reveals an incomplete understanding of layout hierarchies, cell groupings, and formatting cues critical for accurate parsing.

While hallucination (11.26\%) and mathematical errors (9.59\%) are somewhat less frequent, they still suggest inconsistencies in quantitative reasoning and information grounding. Collectively, these findings indicate that despite some robustness in basic vision-language alignment, Qwen2.5-VL requires substantial improvements in handling visually structured data, particularly for tasks requiring deep reasoning and layout-aware comprehension.

\section{Prompts}\label{sec:prompt}
This section contains the exact prompts used in our experiments. 

\begin{tcolorbox}[breakable,colback=lightestgray,colframe=black,title=\centering\textbf{Missing Image Baseline Prompt}]
You will be provided a table in a pipe-separated table where all the entities have been removed. Your task is to:\\\\

\textbf{Step 1: UNDERSTAND THE TABLE CONTEXT} -
Carefully analyze the table structure and identify its purpose and what it mentions.\\\\
\textbf{Step 2: FILL IN THE GAPS} -
Use the table context and your real-world knowledge to deduce the missing entities logically.\\\\
\textbf{Step 3: ANALYZE THE QUESTIONS} -
Read all the questions provided and explore \textit{**ALL TYPES OF REASONING**} to find answers, including but not limited to Numerical reasoning(relationships, totals, and comparisons), Visual reasoning (Colors, shapes, or patterns), Contextual reasoning, (Real-world connections or logic), etc.\\\\

\textbf{Step 4: PROVIDE ANSWERS IN FORMAT} -
Ensure that all answers adhere strictly to the FORMAT specified. Avoid deviating from this format or including unnecessary explanations.

\textbf{\vargreen{\{Answer Formatting Guidelines\}}}

ALWAYS PROVIDE YOUR ANSWERS IN THIS FORMAT.\\

If you are unable to answer it, simple answer \textit{UNKNOWN}.\\\\

I will provide one example to show you:

\textbf{\vargreen{\{One Shot Example\}}}

Now I will provide you with the table and question. 
\\
\textbf{\varred{\{TABLE\}}}
\\
\textbf{\varblue{\{QUESTION\}}}
\\\\
Based on the examples that I have provided and the steps I mentioned above, answer the question.
\end{tcolorbox}

\begin{tcolorbox}[breakable,colback=lightestgray,colframe=black,title=\centering\textbf{Entity Replaced Baseline Prompt}]
You will be provided a pipe-separated table format that contains some entities. Your task is to:\\\\

\textbf{Step 1: UNDERSTAND THE TABLE CONTEXT} -
Carefully analyze the table structure and identify its purpose and what it mentions.\\\\

\textbf{Step 2: ANALYZE THE QUESTIONS} -
Read all the questions provided and explore \textit{**ALL TYPES OF REASONING**} to find answers, including but not limited to Numerical reasoning (relationships, totals, and comparisons), Visual reasoning (colors, shapes, or patterns), Contextual reasoning (real-world connections or logic), etc.\\\\

\textbf{Step 3: PROVIDE ANSWERS IN FORMAT} -
Ensure that all answers adhere strictly to the FORMAT specified. Avoid deviating from this format or including unnecessary explanations.\\
\textbf{\vargreen{\{Answer Formatting Guidelines\}}}

ALWAYS PROVIDE YOUR ANSWERS IN THIS FORMAT.\\\\

\textit{**IMPORTANT**} ALL answers are there in the table. ANALYZE the question and table properly.\\\\

I will provide one example to show you:
\textbf{\vargreen{\{One Shot Example\}}}\\
Now I will provide you with the table and question.\\
\textbf{\varred{\{TABLE\}}}

\textbf{\varblue{\{QUESTION\}}}\\\\

Based on the examples that I have provided and the steps I mentioned above, answer the question.
\end{tcolorbox}

\begin{tcolorbox}[breakable,colback=lightestgray,colframe=black,title=\centering\textbf{Image Captioning Baseline Prompt}]

You will be provided a table in a pipe-separated table with images included. Your task is to:\\\\

\textbf{Step 1: UNDERSTAND THE TABLE CONTEXT} -
Carefully analyze the table structure and identify its purpose and what it mentions.\\\\

\textbf{Step 2: CAPTION EVERY IMAGE} -
Based on the image, provide a caption for that image. Your job is to reason, predict, and replace image entity tags and provide visual descriptions.\\\\

\textbf{Step 3: CREATE A TABLE} -
Based on the image captions, create a pipe-separated table where the image placeholders or cells have been replaced with their captions.\\\\

\textbf{Step 4: ANALYZE THE QUESTIONS} -
Read all the questions provided and explore \textit{**ALL TYPES OF REASONING**} to find answers.\\\\

\textbf{\varred{\{TABLE WITH CAPTIONS\}}}\\\\

\textbf{Step 5: PROVIDE ANSWERS IN FORMAT}
Ensure that all answers adhere strictly to the FORMAT specified.

\textbf{\vargreen{\{Answer Formatting Guidelines\}}}

ALWAYS PROVIDE YOUR ANSWERS IN THIS FORMAT.\\\\

\textit{**IMPORTANT**} ALL answers are there in the table. ANALYZE the question and table properly.\\\\

Now I will provide you with the question.

\textbf{\varblue{\{QUESTION\}}}

Based on the steps I mentioned above, answer the question.
\end{tcolorbox}

\begin{tcolorbox}[breakable,colback=lightestgray,colframe=black,title=\centering\textbf{Table as an Image Baseline Prompt}]
You will be provided an image of a table.
Your task is to:\\\\

\textbf{Step 1: UNDERSTAND THE IMAGE CONTEXT} -
Carefully analyze the image content and understand the tabular structure and all text and visual aspects inside the image.\\\\

\textbf{Step 2: ANALYZE THE QUESTIONS} -
Read all the questions provided and explore \textit{**ALL TYPES OF REASONING**} to find answers.\\\\

\textbf{Step 3: PROVIDE ANSWERS IN FORMAT
}Ensure that all answers adhere strictly to the FORMAT specified. Avoid deviating from this format or including unnecessary explanations.

\textbf{\vargreen{\{Answer Formatting Guidelines\}}}

ALWAYS PROVIDE YOUR ANSWERS IN THIS FORMAT.

\textit{**IMPORTANT**} ALL answers are there in the image.\\\\

Now I will provide you with the image of the table.

\textbf{\varred{\{IMAGE OF TABLE\}}}

For this image, you will answer the following question.

\textbf{\varblue{\{QUESTION\}}}

Based on the steps I mentioned above, answer the question.
\end{tcolorbox}

\begin{tcolorbox}[breakable,colback=lightestgray,colframe=black,title=\centering\textbf{Interleaved Baseline Prompt}]

You will be provided a pipe-seperated table where some cells are images.
Your task is to:\\\\

\textbf{Step 1: UNDERSTAND THE TABLE CONTEXT} -
Carefully analyze the table structure and understand the intricate relationship between image and text.\\\\

\textbf{Step 2: ANALYZE THE QUESTIONS} -
Read all the questions provided and explore **ALL TYPES OF REASONING** to find answers.\\\\

\textbf{Step 3: PROVIDE ANSWERS IN FORMAT
}
Ensure that all answers adhere strictly to the FORMAT specified. Avoid deviating from this format or including unnecessary explanations.

\textbf{\vargreen{\{Answer Formatting Guidelines\}}}

ALWAYS PROVIDE YOUR ANSWERS IN THIS FORMAT.\\\\

\textit{**IMPORTANT**} ALL answers are there in the image.\\\\

Now I will provide you with the table.

\textbf{\varred{\{INTERLEAVED TABLE\}}}

For this table, answer the following question.

\textbf{\varblue{\{QUESTION\}}}

Based on the steps I mentioned above, answer the question.
\end{tcolorbox}

\paragraph{Answer Formatting Guidelines}

The following answer formatting guidelines were provided along with every prompt to eliminate inconsistencies and ensure a uniform response structure across all task types. 
The model is expected to provide only the final answer, formatted as described below:
\vspace{-0.5em}
\begin{itemize}
    \item \textbf{Single Entity:}  
    Return a single \textit{string} representing one entity such as a name, country, company, object, or similar.  
    The answer should be concise and written in one line without extra text.  
    \vspace{-0.75em}
    \begin{itemize}
        \item Example (Name): \texttt{Elon Musk} 
        \vspace{-0.2em}
        \item Example (Country): \texttt{China}
        \vspace{-0.2em}
        \item Example (Company): \texttt{Google} 
        \vspace{-0.2em}
        \item Example (Color): \texttt{Red}
    \end{itemize}
    
    \vspace{-0.5em}
    
    \item \textbf{Single Number:}  
    - If the answer is a whole number, write it \textbf{without decimals}.  
    \vspace{-0.2em}
    - If it has decimals, round to \textbf{two decimal places}.  \vspace{-0.2em}
    - If the last digit after rounding is 0 (e.g., 23.40), remove the trailing zero (→ \textbf{23.4}).  
    Units should only be included if explicitly mentioned in the question. 
    \vspace{-0.75em}
    \begin{itemize}
        \item Example (Whole Number): \texttt{45}  
        \vspace{-0.2em}
        \item Example (Decimal): \texttt{12.36} 
        \vspace{-0.2em}
        \item Example (Trimmed Decimal): \texttt{23.4}
        \vspace{-0.2em}
    \end{itemize}
    \vspace{-0.5em}
    \item \textbf{Multiple Entities:}  
    Provide a \textbf{list of strings}, each following the same rules as the Single Entity format.  
    Use comma-separated values enclosed in square brackets. 
    \vspace{-0.75em}
    \begin{itemize}
        \item Example: \texttt{["Apple", "Microsoft", "Google"]}
    \end{itemize}
    \vspace{-0.75em}
    \item \textbf{Multiple Numbers:}  
    Provide a \textbf{list of numbers}, each following the Single Number formatting rule.  
    Use comma-separated values enclosed in square brackets.  
    \vspace{-0.75em}
    \begin{itemize}
        \item Example: \texttt{[23, 45.67, 89.4]}
    \end{itemize}
    \vspace{-0.75em}
    \item \textbf{Image Locations:}  
    When the answer involves identifying a location within a visual or tabular structure, specify it using the following format:  
    \texttt{row\_num\_col\_num}.  
    \vspace{-0.75em}
    \begin{itemize}
        \item Example: \texttt{row\_2\_col\_3}
    \end{itemize}
\end{itemize}

\end{document}